\documentclass[12pt]{iopart}
\usepackage{color}
\usepackage{tipa}
\usepackage{iopams} 
\usepackage{subfigure}
\usepackage{epsfig}
\usepackage{graphicx}
\usepackage{epstopdf}
\expandafter\let\csname equation*\endcsname\relax
\expandafter\let\csname endequation*\endcsname\relax
\usepackage{amsmath,amssymb}    
\usepackage[numbers]{natbib}
\usepackage[titletoc,toc,title]{appendix}
\usepackage{graphicx}
\usepackage{subfig}
\usepackage{caption}
\usepackage{sidecap}
\usepackage[applemac]{inputenc}
\usepackage{natbib}
\usepackage{graphicx}           
\usepackage{flafter}            
\usepackage{amsmath,amssymb}    
\usepackage{bm}                 
\usepackage{memhfixc}           
\usepackage[activate]{pdfcprot} 
\usepackage{pdfsync}            

\usepackage{verbatim}

\newtheorem{remark}{Remark}

\usepackage{algorithm,algcompatible,amsmath}
\algnewcommand\INPUT{\item[\textbf{Input:}]}%
\algnewcommand\OUTPUT{\item[\textbf{Output:}]}%
\begin{document}
\title{A Dynamical Mean-Field Theory for Learning in Restricted Boltzmann Machines}
\vspace{0.1pc}
\author{Burak \c{C}akmak and Manfred Opper}
\vspace{0.3pc}
\address {Artificial Intelligence Group, Technische Universit\"{a}t Berlin, Germany}
\eads{\mailto{\{burak.cakmak, manfred.opper\}@tu-berlin.de}}

\begin{abstract}
We define a message-passing algorithm for computing magnetizations in Restricted Boltzmann machines, which are Ising models on bipartite graphs introduced as neural network models for probability distributions over spin configurations. To model nontrivial statistical dependencies between the spins' couplings, we assume that the rectangular coupling matrix is drawn from an arbitrary bi-rotation invariant random matrix ensemble. Using the dynamical functional method of statistical mechanics we exactly analyze the dynamics of the algorithm in the large system limit. We prove the global convergence of the algorithm under a stability criterion and compute asymptotic convergence rates showing excellent agreement with numerical simulations.
\end{abstract}





\bibliographystyle{plain}
\def\mathlette#1#2{{\mathchoice{\mbox{#1$\displaystyle #2$}}%
                               {\mbox{#1$\textstyle #2$}}%
                               {\mbox{#1$\scriptstyle #2$}}%
                               {\mbox{#1$\scriptscriptstyle #2$}}}}
\newcommand{\matr}[1]{\mathlette{\boldmath}{#1}}
\newcommand{\RR}{\mathbb{R}}
\newcommand{\CC}{\mathbb{C}}
\newcommand{\NN}{\mathbb{N}}
\newcommand{\ZZ}{\mathbb{Z}}
\newcommand{\at}[2][]{#1|_{#2}}
\def\oneh{\frac{1}{2}}
\newcommand{\Kb}{{\mathbf{K}}}
\section{Introduction}
In recent years there has been a renewed interest in the application of statistical mechanics 
ideas to the study of large neural networks and other related learning models \cite{gardner1988space,watkin1993statistical,opper1996statistical,nishimori2001statistical,mezard2009information}.
While earlier research in the field concentrated on static properties of such models, a major focus is now on the understanding of the dynamics
of message-passing algorithms for probabilistic data models.  Such algorithms, under certain statistical assumptions on network couplings,  provide
efficient and accurate computations for averages of probabilistic network nodes in the large system limit. Current research concentrates on
models with dense connectivities and the corresponding AMP (approximate message-passing) and VAMP algorithms \cite{Bolthausen,Bayati,Opper16,cakmak17,rangan2019vector,takeuchi,fletcher2018inference,ccakmak2020analysis}. Fixed points of these algorithms are known to be solutions of the static TAP (Thouless--Anderson--Palmer) mean-field equations for the expectations of nodes \cite{Adatap,Minka1,OW5}. The models studied so far can usually be described in terms of non-Gaussian probabilistic nodes which are coupled by pairwise random interactions. This will include Ising models (aka ``Boltzmann machines'' in the machine learning community),
but also Bayesian classifiers and models of sparse signal recovery. 

Less work, from a statistical mechanics perspective, has been devoted to a conceptually simple machine learning model of the Ising type, the so-called {\em restricted Boltzmann machine} (RBM) first introduced by \cite{smolensky1986information} and later studied extensively by J Hinton \cite{hinton2007boltzmann}. The Ising model is defined by a bipartite graph of spins---the ``neurons''---which belong to either visible or hidden ones and there are only connections between neurons of different groups. This model can learn a probability distribution over visible spin configurations by adjusting couplings between the spins. Training of an RBM aims at increasing the probability of observations from the visible neurons for a set of data. Gradients of this likelihood can be expressed as in terms of conditional moments of the hidden spins given the observed visible ones and of moments for the joint distribution of both visible and hidden units based on the RBM model.

While the conditional moments are easily computed analytically, the exact computation of the model moments becomes intractable for large systems. Hence, other methods which approximate the gradients such as the
{\em contrastive divergence algorithm} \cite{2002training,tieleman2008training} are used in practice. From the point of view of statistical mechanics however, a direct approximation of intractable statistical averages by message-passing methods seems to be a sensible alternative. For applications of this technique see \cite{Minka1,OW5,baker2020tramp}. 

A first approach to develop a theoretical background for such a method
is to study the thermodynamic properties of the RBM for quenched independent random couplings. This has been done in the recent papers \cite{decelle2018thermodynamics,hartnett2018replica}.  On the other hand,  iterative learning algorithms for adapting the couplings to data will introduce statistical dependencies between the couplings. Hence, an extension of the theory which allows for dependencies would be desirable. Finally, the development of an AMP style message-passing algorithm, which can be analyzed exactly in the thermodynamic limit, would be necessary.

In this paper, we will present a step in this direction. The main novel contributions of our paper are:  We consider the statistical mechanics of RBMs with couplings from bi-rotation invariant random matrix ensembles
which allow for weak dependencies. The static properties of the model are computed by the replica method and TAP equations for the bi-rotation invariant case are derived.
We then construct an AMP style algorithm which has the solutions of the TAP equations
as fixed points. The algorithm is made efficient by utilizing order parameters computed from the replica result. Finally, we analyze the dynamics of the algorithm in the large system
limit using {\em dynamical functional theory} (DFT) generalizing our previous papers \cite{Opper16,CakmakOpper19}. The quenched averages over the ensemble of coupling matrices require nontrivial extensions of the previously developed analytical techniques. We give a proof of convergence of the algorithm from random initial conditions and compute convergence rates analytically.

The paper is organized as follows: In Section~II we introduce the Ising model for the RBMs and also briefly present the learning problem of RBMs. Section~III presents the replica-symmetry (RS) calculation of the free energy and the TAP equations of the magnetizations for general bi-rotation invariant random coupling matrix ensembles. In Section~IV we present our new algorithm for solving the TAP equations and in Section~V we present its DFT analysis. Section~VI provides convergence properties of the algorithm.  In Section~VII we present algorithmic considerations to compute model parameters that are needed by the algorithm before the iteration starts. Comparisons of the theory with simulations are given in Section~IIX. Section~IX presents a summary and outlook. The derivations of our results are located in the Appendix.

\section{Ising models for restricted Boltzmann machines}
We consider Ising models where the joint distribution of the vectors of spins $\matr s_1\in \{-1,+1\}^{N_1\times 1}$ and $\matr s_2\in \{-1,+1\}^{N_2\times 1}$ is given by the (conditional) Gibbs-Boltzmann distribution 
\begin{equation}
p(\matr s_1,\matr s_2\vert \matr W,\matr h_1,\matr h_2)\doteq \frac{1}{Z}
\exp\left(\matr s_1^\top\matr W\matr s_2+\matr s_1^\top\matr h_1+\matr s_2^\top\matr h_2\right)\label{Gibbs2}
\end{equation}
with $Z$ denoting the normalization constant.  

\subsection{Motivation: Learning of restricted Boltzmann machines}
Consider a dataset $\mathcal D\doteq\{\matr s_1^{(1)},\matr s_1^{(2)},\cdots,\matr s_1^{(D)}\}$ whose elements $\matr s_1^{(d)}\in \{-1,1\}^{N_1\times 1}$ are assumed to be drawn independently from a generative distribution
\begin{equation}
p(\matr s_1\vert \matr W,\matr h_1,\matr h_2)\doteq\sum_{\matr s_2}p(\matr s_1,\matr s_2\vert \matr W,\matr h_1,\matr h_2).
\end{equation}
Here, the vector $\matr s_2$ stands for the vector of hidden (i.e. unobservable) units. 
The learning problem of RBMs is to perform the maximum-likelihood estimations of the model parameters $\{\matr W, \matr h_1,\matr h_2 \}$. The learning problem could be performed by using gradient descent which requires the computations of the gradients of the likelihood as
\begin{align}
\nabla W_{ij}\ln p(\mathcal D\vert \matr W, \matr h_1,\matr h_2)&\propto\frac 1 D\sum_{d\leq D}\mathbb \langle s_{1i}^{(d)}s_{2j}\rangle-\mathbb E[s_{1i}s_{2j}]\\
\nabla h_{1i}\ln p(\mathcal D\vert \matr W, \matr h_1,\matr h_2)&\propto\frac 1 D\sum_{d\leq D}s_{1i}^{(d)}-\mathbb E[s_{1i}]\\
\nabla h_{2j}\ln p(\mathcal D\vert \matr W, \matr h_1,\matr h_2)&\propto\langle s_{2j}\rangle-\mathbb E[s_{2j}].
\end{align}
Here, $\mathbb E[\cdot]$ and $\langle \cdot  \rangle $ stand for the expectations over the Gibbs-Boltzmann distribution \eqref{Gibbs2} and the distribution $p(\matr s_2\vert\matr s_1^{(d)},\matr W,\matr h_1,\matr h_2)$, i.e. \emph{model}  and \emph{clamped} expectations, respectively.  Evidently, exact computations of the model expectations are impractical for large systems. 
On the other hand, the clamped expectations involve factorizing distributions only.
A Monte Carlo method to approximate these expectations \cite{2002training,tieleman2008training} could be problematic for large systems.

Motivated by the recent study \cite{Tramel18} we consider a TAP-based approach for computing the model expectations of the spin variables (the magnetizations). Thanks to the linear response relation $\mathbb E[s_{1i}s_{2j}]=\frac{\partial\mathbb E[s_{1i}]}{\partial h_{2j}}+\mathbb E[s_{1i}]\mathbb E[s_{2j}]$, the problem of computing model expectations in the parametric approach reduces to the computation of the magnetizations, solely.

\section{General bi-rotation invariant random matrix ensembles}
For the sake of simplicity of analysis, we will limit our attention to the case of the identical
 ``external-fields''
\begin{equation}
h_{1i}=h_1\neq 0 \quad \text{and} \quad  h_{2j}=h_2\neq 0\qquad \forall i,j.
\end{equation}
Moreover, in order to allow for nontrivial dependencies between couplings elements $\{W_{ij}\}$, we assume that the coupling matrix is drawn from an arbitrary \emph{bi-rotation invariant} random matrix ensemble. Specifically, the (probability) distribution of the coupling matrix $\matr W$ is invariant under multiplications from both left and right with any independent orthogonal matrices \cite{livan2018introduction}. Equivalent, we have the spectral decomposition \cite{collins2014integration}
\begin{equation}
\matr W=\matr O\matr \Sigma \matr V^\top
\end{equation}
where the matrices in the product are mutually independent and $\matr O\in \RR^{N_1\times N_1}$ and $\matr V\in \RR^{N_2\times N_2}$ are Haar (random) orthogonal matrices. This choice of an ensemble is rich enough to allow for a free choice of singular values of matrices, but it considers that (left and right) eigenvectors are in ``general position''. 
\subsection{Rectangular Spherical Integration}
Previous statistical mechanics analyses \cite{Opper16,CakmakOpper19} involve \emph{symmetric} random matrices and usage of the asymptotic Itzykson-Zuber integration \cite{Zuber,Collins5} in the analyses becomes useful. On the other hand, we now need to sort out the analysis involving the non-symmetric (and rectangular, in general) random matrix $\matr W$ and it is not clear how to use the Itzykson-Zuber integral within this context. It turns out that the method of (asymptotic) ``\emph{rectangular spherical integration}'' \cite{Kab08,Benaych-Georges2011,maillard2019high} becomes an appropriate approach within current context.

Specifically, for an $N_2\times N_1$ matrix $\matr Q$ independent of $\matr W$ we write \cite[Section 5.5.1]{maillard2019high}
\begin{equation}
\lim_{N_1\to \infty}\frac{1}{N_1}\ln \mathbb E_{\matr O,\matr V}[{\mathrm e}^{\sqrt{N_1N_2}{\rm tr}(\matr Q\matr W)}]= \frac{1}{2}{\rm tr}(I(\matr Q\matr  Q^\top))
\end{equation}
where we have defined the generating function
\begin{equation}
{I}(x)\doteq \sup_{\psi_1,\psi_2}\left\{\psi_1+\alpha\psi_2+(1-\alpha)\ln \psi_2-\int {\rm dP}_{\matr W}(t)\; \ln(\psi_1\psi_2-xt) \right\}- (1+\alpha).\label{Ix}
\end{equation}
Here, ${\rm P}_{\matr W}$ stands for the limiting spectral distribution of the Gramian $\matr W\matr W^\top$ and we introduce the aspect ratio $\alpha\doteq N_2/N_1$ which is assumed to be fixed as $N_1,N_2\to \infty$. 

Next we give some specific examples of the generating function $I(x)$ for the random matrix ensembles from which we shall exemplify our general arguments:
\begin{itemize}
	\item [(i)] (i.i.d. random couplings) $\matr W$ has independent (Gaussian) entries with zero mean and variance $\beta/N_1$. In this case, we have
	\begin{equation}
	I(x)=\alpha\beta x. \label{iid}
	\end{equation}
	\item [(ii)](Column-orthogonal random coupling matrices)  $\matr W$ has random orthogonal columns as $\matr W=\sqrt{\beta}\matr O\matr P_\alpha$ where $\matr O$ is Haar random orthogonal and $\matr P_\alpha$ is the $N_1\times N_2$ rectangular projection matrix with the entries $(\matr P_\alpha)_{ij}=\delta_{ij}$. In this case,  we have
\begin{equation}
I(x)=\sqrt{1+4\alpha\beta x}-\ln(1+\sqrt{1+4\alpha\beta x})+\ln 2-1.
\end{equation} 
\end{itemize}
As regards to the model (ii) we note, that in the context of RBM the number of visible variables is typically larger than the number of hidden variables, i.e. $N_1\geq N_2$. Therefore, we do not address the row-orthogonal case. Yet, by symmetry it can treated similarly. 
\subsection{Replica-symmetry calculation of the free energy and the static order parameters}
Using the rectangular spherical integration we perform the RS calculation of 
the log--partition function in \ref{rep}. The result is given by
\begin{align}
\frac{1}{N_1}\mathbb E \ln Z&\simeq \operatorname*{extr}_{\{\chi_k,\hat q_k\}}\left\{\mathbb E[\ln 2\cosh(h_1+\sqrt{\hat q_1}u)]+\alpha  \mathbb E[\ln 2\cosh(h_2+\sqrt{\hat q_2}u)]\right.+\nonumber \\
&\left.\quad \qquad-\frac{1}{2}(\chi_1\hat q_1+\alpha\chi_2{\hat q_2}) +\frac 1 2 I(\chi_1\chi_2)-\frac{1}{2}(\chi_1+\chi_2+2\chi_1\chi_2)I'(\chi_1\chi_2)\right\} \label{free}
\end{align}
where $I'$ stands for the derivative of $I$ and the random variable $u$ is a standard (zero mean, unit variance) normal Gaussian. Furthermore, extremizations of \eqref{free} with respect to the order parameters $\{\chi_k,\hat q_k\}$ give the fixed-point equations of the order parameters as
\begin{subequations}
\label{s.order}
\begin{align}
\chi_1&=\mathbb E[\tanh'(h_1+ \sqrt{\hat q_1}u)]\\
\chi_2&=\mathbb E[\tanh'(h_2+\sqrt{\hat q_2}u)]\\
\hat q_1&=\chi_2^2(1-\chi_1){I}''(\chi)+(1-\chi_2)(I'(\chi)+\chi I''(\chi))\\
\hat q_2&=\frac {\chi_1^2(1-\chi_2){I}''(\chi)+(1-\chi_1)(I'(\chi)+\chi I''(\chi))}{\alpha}
\end{align}
\end{subequations}
where we have defined $\chi\doteq \chi_1\chi_2$. For example, in the case of the i.i.d. random couplings we get from \eqref{iid}
\begin{equation}
	\hat q_1 =(1-\chi_2)\alpha\beta \quad \text{and}\quad \hat q_2=(1-\chi_1)\beta
\end{equation}
and the resulting free energy agrees with the previous RS calculations~{\cite{decelle2018thermodynamics,hartnett2018replica}.}

When the analytical expressions of $I'(\chi)$ and $I''(\chi)$ are not available, we can consider a practical
approach for computing them for a given empirical spectral distribution of~$\matr W\matr W^\top$. For details, we refer the reader to  Section~\ref{comp_order}. 

\subsection{TAP Equations}
Using a cavity method \cite{Mezard} along with arguments from asymptotic freeness properties of random matrices \cite{Hiai} we derive in \ref{dervtap} the TAP (fixed-point) equations of the magnetizations (specifically $\matr m_k\doteq \mathbb E[\matr s_k]$ for $k=1,2$). They are given by
\begin{subequations}
	\label{tap2}
	\begin{align}
	\matr m_1&=\tanh(h_1+\matr \gamma_1)\\
	\matr m_2&=\tanh(h_2+\matr \gamma_2)\\
	\matr \gamma_1&=\matr W\matr m_2-\chi_2I'(\chi)\matr m_1\\
	\matr \gamma_2&=\matr W^\top\matr m_1-\frac{\chi_1I'(\chi)}{\alpha}\matr m_2
	\end{align}	
\end{subequations}
where $\chi=\chi_1\chi_2$ and $\{\chi_k\}$ are solutions of the equations \eqref{s.order}. For example, in the case of the i.i.d. random couplings we have $I'(\chi)=\alpha\beta$, so that the TAP equations read as 
\begin{subequations}
	\label{iidstad}
	\begin{align}
	\matr m_1&=\tanh(h_1+\matr W\matr m_2-\alpha\beta\chi_2\matr m_1)\\
	\matr m_2&=\tanh(h_2+\matr W^\top\matr m_1-\beta\chi_1\matr m_2).
	\end{align}	
\end{subequations}
The equations \eqref{iidstad} are consistent with those derived in \cite{Tramel18} using a high temperature expansion approach of the free energy. 

\subsection{The spin cross-correlations}
We next address  the TAP equations for computing the spin cross-correlations, e.g. $\mathbb E[s_{1i}s_{2j}]$. To this end,  
we introduce the spin covariance matrix as
\begin{equation}
\matr \chi\doteq \mathbb E[\matr s\matr s^\top]-\mathbb E[\matr s]\mathbb E[\matr s]^\top\quad \text{with} \quad \matr s\doteq {\small\left[\begin{array}{c}
	\matr s_1\\
	\matr s_2
	\end{array}\right]}.
\end{equation}
By linear response the TAP equations \eqref{tap2} yields the (approximate) covariance matrix as
\begin{equation}
\matr \chi= \left(\begin{array}{cc}
\matr \Lambda_1&-\matr W\\
-\matr W^\top& \matr\Lambda_2
\end{array}\right)^{-1}   \label{covariance}
\end{equation}
where we have introduced the diagonal matrices $\matr \Lambda_1$ and $\matr \Lambda_2$ with the diagonal entries
\begin{align}
(\matr \Lambda_1)_{ii}&=\frac{1}{\tanh'(h_{1}+\gamma_{1i})}+\chi_2I'(\chi)\\
(\matr \Lambda_2)_{jj}&=\frac{1}{\tanh'(h_{2}+\gamma_{2j})}+\frac{\chi_1I'(\chi)}{\alpha}.
\end{align}
In particular, from \eqref{covariance} we have the (approximate) cross-correlations 
\begin{equation}
\mathbb E[s_{1i}s_{2j}]=(\matr \Lambda_1^{-1}\matr W(\matr \Lambda_2-\matr W^\top \matr\Lambda_1^{-1}\matr W)^{-1})_{ij}+\tanh(h_{1}+\gamma_{1i})\tanh(h_{2}+\gamma_{2j}).
\end{equation}
\subsection{Stability of the TAP equations}
A cruial argument in deriving TAP equations is the assumption of weak dependencies between the spins \cite{Mezard}. Specifically, the  off-diagonal entries of the spin covariance matrix $\matr \chi$ should vanish as $O(1/\sqrt{N_1})$ for $N_1,N_2\to\infty$ (with the ratio $\alpha=N_2/N_1$ fixed). We sort out the consistency of the weak-dependencies assumption by studying the condition
\begin{equation}
\mathbb E[(\chi_{nn'})^2]=O(\frac{1}{N_1})\quad \forall n\neq n' \label{s2}
\end{equation}
where $\matr \chi $ is given by \eqref{covariance} and the expectation is taken over the random matrix $\matr W$. This condition implies the convergence $\chi_{nn'}\to 0,\forall n\neq n'$ in a $L^2$ norm sense. We show in \ref{derstab} that the condition \eqref{s2} is fulfilled if and only if the following bounds hold
\begin{equation}
{\rm R}_k'\mathbb E[(\tanh'(h_k+\sqrt{\hat q_k}u))^2]<1\quad k=1,2.\label{AT} \\
\end{equation} 
Here, we have defined 
\begin{subequations}
	\label{ATr}
	\begin{align}
	{\rm R}_1'&\doteq\frac{\chi_2}{\chi_1}\left[ \frac{(\alpha+\chi I'(\chi))(I'(\chi)+\chi I''(\chi))}{\alpha-\chi^2 I''(\chi)}-I'(\chi)\right]\\
	{\rm R}_2'&\doteq \frac{\chi_1}{\chi_2}\left[\frac{(1+\chi I'(\chi))(I'(\chi)+\chi I''(\chi))}{\alpha-\alpha\chi^2 I''(\chi)}-\frac{I'(\chi)}{\alpha}\right].
	\end{align}
\end{subequations}
For example, in the case of the i.i.d. random couplings, ${\rm R}_1=\alpha\beta^2\chi_{2}^2$ and ${\rm R}_2=\alpha\beta^2\chi_{1}^2$.

\section{Iterative solution of the TAP equations}
We are looking for a solution to the TAP equations \eqref{tap2} in
terms of iterations of a vector of auxiliary variables $\matr \gamma(t)\doteq{\small \left[\begin{array}{c}
	\matr \gamma_1(t)\\
	\matr \gamma_2(t)
	\end{array}\right]}$,
where $t=1,2\ldots$ denotes the discrete-time index of the iteration. To this end, we will introduce a VAMP-style iterative algorithm \cite{rangan2019vector,takeuchi,Ma,minka2005divergence}. The conventional VAMP approach leads to an iterative algorithm requiring the computation of products of $N_k\times N_k$ matrices for updating certain order parameters at every iteration step, see \cite[Algorithm~3]{Tramel18}. This could be problematic for large $N_k$ and large times. On the other hand, we will devise a VAMP-style algorithm that makes use of the static the order parameters in the RS calculation \eqref{s.order}. This approach allows us to bypass the need for products of large matrices. Specifically, we propose the following iterative algorithm
	\label{newdynamics}
	\begin{align}
	\matr \gamma(t)&=\matr Af(\matr \gamma(t-1)) \quad \text{with} \quad \matr \gamma(0)={ \left[\begin{array}{c}
			\sqrt{\hat q_1}\matr u_1\\
		\sqrt{\hat q_2}\matr u_2
			\end{array}\right]}
	\end{align}
which is solely based on \emph{matrix vector} multiplications and evaluations of a scalar nonlinear function $f$. Here, the entries of the vectors $\matr u_1\in\RR^{N_1\times 1}$ and $\matr u_2\in \RR^{N_2\times 1}$ are drawn independently from a normal Gaussian distribution. Furthermore, for a vector $\matr x\doteq{\small \left[\begin{array}{c}
	\matr x_1\\
	\matr x_2
	\end{array}\right]}$ with $\matr x_k\in{\RR}^{N_k\times 1}$ we have introduced the function
\begin{equation}
f(\matr x)\doteq{ \left[\begin{array}{c}
		f_1(\matr x_1)\\
		f_2(\matr x_2)
	\end{array}\right]} \quad \text{with}\quad f_k(x)\doteq \frac{1}{\chi_k}\tanh(h_k+x)-x.
\end{equation} 
Moreover, we define the \emph{time-independent} matrix $\matr A$ as
\begin{equation}
\matr A\doteq\left(\begin{array}{cc}
\psi_1{\bf I} &-\chi_2\matr W\\
-\chi_1\matr W^\top&\psi_2{\bf I}
\end{array}\right)^{-1}-{\bf I}
\end{equation} 
where  we have introduced the scalars
\begin{align}
\psi_1\doteq 1+\chi I'(\chi) \quad \text{and}\quad   \psi_2\doteq 1+\frac{\chi I'(\chi)}{\alpha}. \label{psi}
\end{align}
Actually,  the variables $\psi_1$ and $\psi_2$ are those extremizing $I(\chi)$ in \eqref{Ix}, see \ref{Apre}. 

It is easy to show that the fixed points of $\matr\gamma(t)$ coincide with the solution of the TAP equations for $\matr \gamma\doteq{\small \left[\begin{array}{c}
	\matr \gamma_1\\
	\matr \gamma_2
	\end{array}\right]}$, if we identify the
corresponding vectors of magnetizations by
\begin{equation}
\matr m_k={\chi_k}(\matr \gamma_k+f_k(\matr \gamma_k)). 
\end{equation} 
 
\section{The dynamical functional analysis}
In this section, we analyze the dynamical properties of the iterative algorithm using the method of the dynamical functional analysis \cite{Martin,Eisfeller,CakmakOpper19}. Our goal is deduce the statistical properties of marginals $\{\gamma_{1i}(t)\}\doteq\{\gamma_{1i}(t)\}_{0\leq t\leq T}$ and $\{\gamma_{2j}(t)\}$. To this end, we introduce the moment \emph{generating-functional} for the trajectories of  $\{\gamma_{1i}(t)\}$ and $\{\gamma_{2j}(t)\}$ as 
\begin{align}
Z_{ij}(\{l_{1}(t),l_{2}(t)\})\doteq\int \prod_{t=1}^T &  {\rm d}\matr \gamma(t)\; 
\delta\left[\matr \gamma(t)-\matr A{f}(\matr \gamma(t-1))\right]{\mathrm e}^{{\rm i}\sum_{t=0}^{T}[l_{1}(t)\gamma_{1i}(t)+l_{2}(t)\gamma_{2j}(t)]}. \label{gff}
\end{align}
We are interested in computing the averaged generating functional $\mathbb E[Z_{ij}(\{l_{1}(t),l_{2}(t)\})]$ where the expectation is taken over the Haar random matrices $\matr O$ and $\matr V$ and the random initialization $\matr\gamma(0)$. From the averaged generating functional, e.g., we may compute 
\begin{align}
\frac{1}{N_k}\mathbb E[\matr \gamma_k(t)^\top\matr\gamma_k(s)]&=\mathbb E[\gamma_{ki}(t)\gamma_{ki}(s)] \label{ex}\\
&=-\left.\frac{\partial\mathbb E[Z_{ij}(\{l_{1}(t),l_{2}(t)\})]}{\partial l_k(t)\partial l_k(s)}\right|_{\{l_1(t),l_2(t)\}=0}. \label{dfmse}
\end{align}
Using \eqref{dfmse} we can quantify the averaged-normalized-square Euclidean distance between iterates of the algorithm at different times (i.e. $\frac{1}{N_k}\mathbb E[\Vert\matr \gamma_k(t)-\matr \gamma_k(s)\Vert^2]$) which will allow us to analyze the convergence properties of the dynamics.
We defer the explicit and lengthy computation of the DFT analysis to \ref{derDF}. There, we show that 
\begin{equation}
\mathbb E[Z_{ij}(\{l_1(t),l_2(t)\})]\simeq Z_1(\{l_1(t)\})\times Z_2(\{l_2(t)\})
\end{equation}
where (for $k=1,2$) we have defined the single-site generating functionals
\begin{align}
 Z_k(\{l(t)\})\doteq \int \{{\rm d}\gamma_k(t)\}\; \mathcal N(\gamma_k(0)\vert 0,\hat q_k)\mathcal N(\gamma_k(1),\dots,\gamma_k(T)\vert \matr 0, \mathcal C_{\gamma_k})\;{\mathrm e}^{ {\rm i}\sum_{t=0}^{T}\gamma_k(t)l(t)}\label{sef}
\end{align}
Here, $\mathcal N(\cdot \vert \matr \mu,\matr \Sigma)$ denotes the Gaussian density function with mean $\matr \mu$ and covariance~$\matr\Sigma$. Thus, we have obtained the ``effective'' stochastic processes for the dynamics of single, arbitrary
components $\gamma_k(t)$ of the vectors $\matr \gamma_k(t)$ such that
\begin{equation}
(\gamma_k(1),\dots,\gamma_k(T))\sim \mathcal N(\matr 0,\mathcal C_{\gamma_k})
\end{equation}
and $\gamma_k(0)\sim \mathcal N(0,\hat q_k)$ is independent of $\{\gamma_k(t)\}_{t\geq 1}$.  Here, the $T\times T$ covariance matrices $\mathcal C_{\gamma_1}$ and $\mathcal C_{\gamma_2}$ are computed by the recursion 
\begin{align}
\left[\begin{array}{c}
\mathcal C_{\gamma_1}(t,s)\\\mathcal C_{\gamma_2}(t,s)
\end{array}\right]= \left[\begin{array}{cc}
a_{11}&a_{12}\\
a_{21}&a_{22}
\end{array}\right] 
\left[\begin{array}{c}
\mathbb E[f_{1}(\gamma_1(t-1))f_{1}(\gamma_1(s-1))]\\
\mathbb E[f_{2}(\gamma_2(t-1))f_{2}(\gamma_2(s-1))]
\end{array}\right]\label{covex1}.
\end{align}
The coefficients $\{a_{kk'}\}$ can be explicitly expressed in terms of $I'(\chi)$ and $I''(\chi)$, see \eqref{showfgood}. Actually, we show in \ref{dcovex2} that these coefficients coincide with the limits
\begin{align}
a_{kk'}&=\lim_{N_k\to \infty} \frac{1}{N_k}{\rm tr}(\matr A_{kk'}\matr A_{kk'}^\top) \quad \text{with}\quad \matr A=\left[\begin{array}{cc}
\matr A_{11} &\matr A_{12} \\
\matr A_{21} &\matr A_{22} 
\end{array}\right]\label{dergf}
\end{align}
where $\matr A_{11}$ has dimension $N_1\times N_{1}$. 

\section{Convergence of the single-variables dynamics}
We analyze the thermodynamic convergence properties of the sequence $\matr \gamma_k(t)$ (for $k=1,2$)  by studying the deviation between the dynamical variables at different times 
\begin{align}
\Delta_{\gamma_k}(t,s)&\doteq \lim_{N\to\infty}\frac{1}{N} \mathbb E[\Vert \matr \gamma_k(t)-\matr \gamma_k(s) \Vert^2]\\
&=\mathcal C_{\gamma_k}(t,t)+\mathcal C_{\gamma_k}(s,s)-2\mathcal C_{\gamma_k}(t,s)\\
&= 2(\hat q_k-\mathcal C_{\gamma_k}(t,s))\label{nice}
\end{align}
Here, the equation \eqref{nice} follows from the fact, by the definition of the recursion \eqref{covex1} we have that (see \ref{variance terms})
\begin{equation}
\mathcal C_{\gamma_k}(t,t)=\hat q_k,\forall t.
\end{equation} 

The two-time covariances have the strictly increasing property
\begin{equation}
\mathcal C_{\gamma_k}(t-1,s-1)<\mathcal C_{\gamma_k}(t,s)<\hat q_k,\quad \forall t\neq s.\label{strictp}
\end{equation}
Furthermore, they converge to the limits
\begin{equation}
\lim_{t,s\to \infty}\mathcal C_{\gamma_k}(t,s)=\hat q_k
\end{equation}
if and only if the following condition holds 
\begin{equation}
\mu_\gamma\doteq\frac{1}{2}(g'_1a_{11}+g'_2a_{22})+\frac 1 2\sqrt{(g'_1a_{11}-g'_2a_{22})^2+4g'_1g'_2a_{12}a_{21}}<1\label{newbound}
\end{equation}
with $g'_k\doteq \mathbb E[(f_k'(h_k+\sqrt{\hat q_k}u))^2]$. Moreover, the rates of convergence to  these limits are the same and given by
\begin{equation}
\lim_{t,s\to\infty}\frac{\Delta_{\gamma_k}(t+1,s+1)}{\Delta_{\gamma_k}(t,s)}=\mu_\gamma,\quad k=1,2.
\end{equation} 
The dynamical stability $\mu_\gamma<1$ ensures the stability of the TAP equations \eqref{AT}. The derivations of these results are given in \ref{dervconv}.

\section{Algorithmic consideration}\label{comp_order}
In this section, we will introduce an algorithmic simplification which bypasses the need for analytical expressions of $I'(\chi)$ and $I''(\chi)$ for computing the necessary order parameters. The approach is based on expressing the order parameters via the (limiting) Green function
\begin{equation}
{\rm G}_\matr W(z)\doteq \lim_{N_1\to \infty} \frac{1}{N_1}{\rm tr}(({z}{\bf I}-\matr W\matr W^\top)^{-1}).\label{Gf}
\end{equation}
Specifically, we show in \ref{dcovex2} that the fixed-point equations of the necessary order parameters $\{\chi_k,\hat q_k,\psi_k\}$ (see \eqref{s.order} and \eqref{psi}) can be equivalently expressed as 
\begin{subequations}
\label{newfp}
\begin{align}
\left[\begin{array}{c}
\chi_1\\ \chi_2
\end{array}\right]&=\left[\begin{array}{c}
\mathbb E[\tanh'(h_1+\sqrt{\hat q_1}u)]\\ \mathbb E[\tanh'(h_2+\sqrt{\hat q_2}u)]
\end{array}\right]\\
\left[\begin{array}{c}
\psi_1\\ \psi_2
\end{array}\right]&=\left[\begin{array}{c}
\frac{\chi}{{\rm G}_{\matr W^\top}(\lambda)}\\ \frac{\chi}{{\rm G}_\matr W(\lambda)}
\end{array}\right]\\
\matr\Theta&=-\left[\begin{array}{cc}
\frac{\psi_2^2}{\chi^2}{{\rm G}}_\matr W'(\lambda)+1
&  \frac{1}{\chi_1^2}(\lambda{\rm G}'_\matr W(\lambda)+{\rm G}_\matr W(\lambda))
\\
\frac{1}{\alpha\chi_2^2}(\lambda{\rm G}'_\matr W(\lambda)+{\rm G}_\matr W(\lambda))&\frac{\psi_1^2}{\chi^2}{{\rm G}}'_{\matr W^\top}(\lambda)+1
\end{array}\right]\\
\left[\begin{array}{c}
\hat q_1\\ \hat q_2
\end{array}\right]&= ({\bf I}+\matr \Theta)^{-1}\matr \Theta\left[\begin{array}{c}
\frac{1-\chi_1}{\chi_1^2}
\\ \frac{1-\chi_2}{\chi_2^2}
\end{array}\right]
\end{align}	
\end{subequations}
with noting that $\lambda=\frac{\psi_1\psi_2}{\chi}$ and $\chi=\chi_1\chi_2$. Hence, the necessary order parameters can be obtained by iteratively solving the equations \eqref{newfp} which require the analytical expressions of the Green functions (i.e. ${\rm G}_\matr W$ and  ${\rm G}_\matr W^\top$) and their derivatives (i.e. ${\rm G}'_{\matr W}$ and  ${\rm G}'_{\matr W^\top}$). Here, we note from \eqref{Gf} the general relations \cite{debbah}
\begin{align}
{\rm G}_{\matr W}(z)&=\alpha{\rm G}_{\matr W^\top}(z)+\frac{1-\alpha}{z} \\
{\rm G}'_{\matr W}(z)&=\alpha{\rm G}'_{\matr W^\top}(z)-\frac{1-\alpha}{z^2}.
\end{align} 
For a practical application of the algorithm 
we can simply approximate the Green function for the Gramian $\matr W^\top \matr W$ and its derivative with their finite-size approximations as
\begin{align}
{\rm G}_{\matr W^\top}(z)\simeq\frac{1}{N_2}\sum_{j\leq N_2} \frac{1}{z-d_j}\quad \text{and}\quad {\rm G}_{\matr W^\top}'(z)\simeq -\frac{1}{N_2}\sum_{j\leq N_2} \frac{1}{(z-d_j)^2}
\end{align}
where $\{d_j\}$ are the eigenvalues of the Gramian $\matr W^\top \matr W$. 
\section{Simulation results}
In this section, we compare our analytical results  with simulations of the algorithm for both random matrix models (i) and (ii).  The simulation results are based on \emph{single instances} of large random matrices $\matr W$. 

\begin{figure}
	\centering 
	\epsfig{file=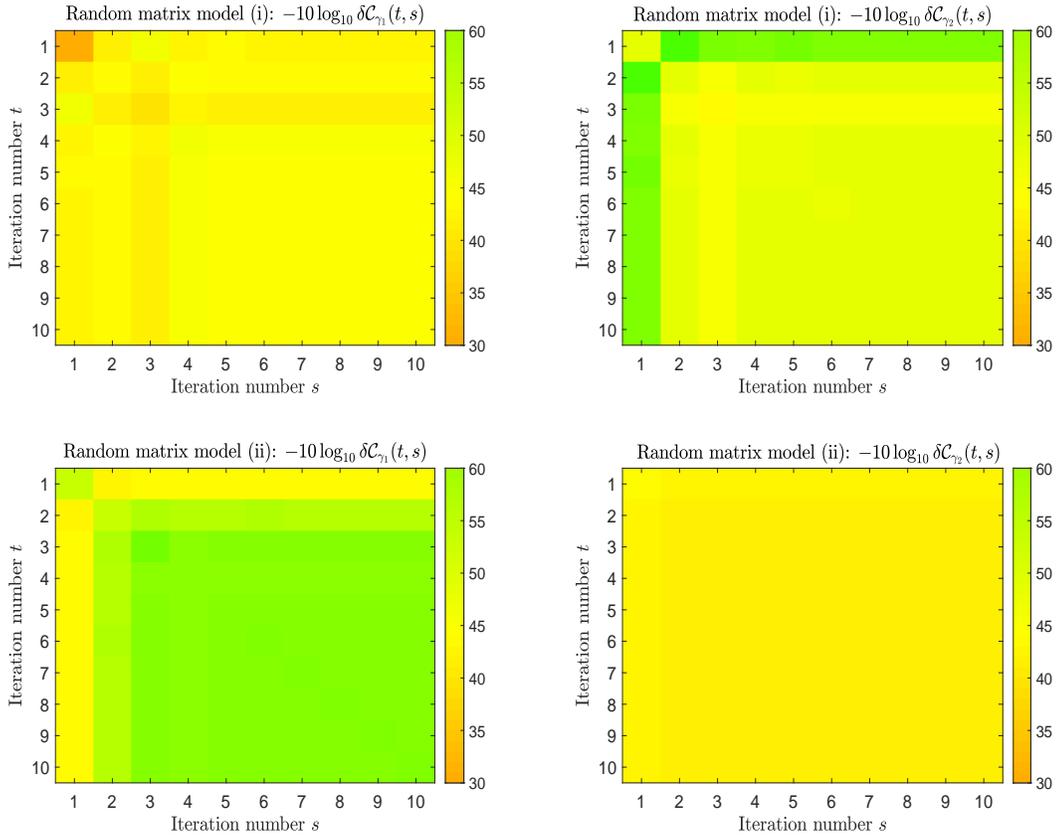,width=1.06\columnwidth,height=12.3cm}
	\caption{Discrepancy between theory and simulations for the 
		two-time covariances with $N_1=10^{4}$, $N_2=N_1/2$, $h_1=2$, $h_2=1$ and $\beta=2$.}\label{fig1}
\end{figure}

\begin{figure}
	\centering
	\epsfig{file=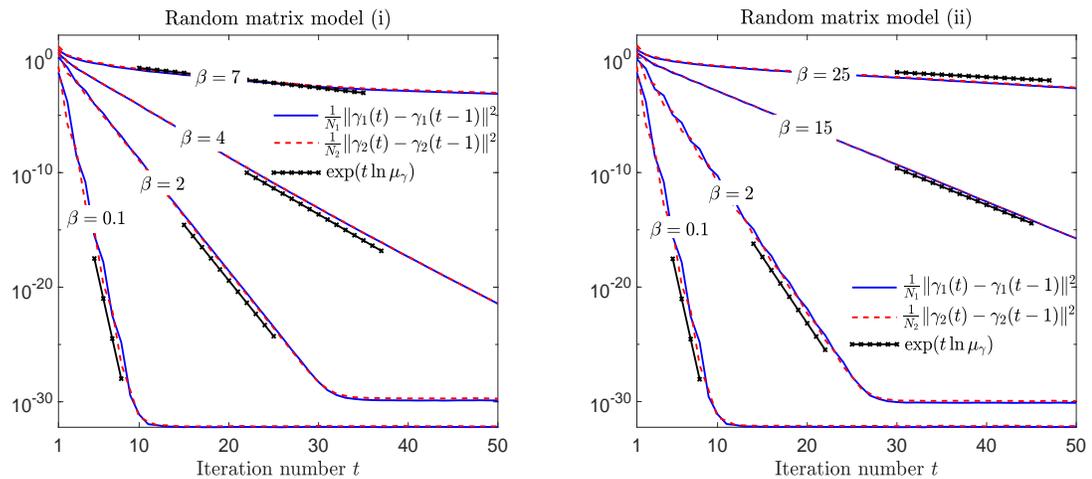,width=1.08\columnwidth,height=6.4cm}
	\caption{Asymptotics of the algorithm with $N_1=10^{4}$, $N_2=N_1/2$, $h_1=2$, $h_2=1$. The ﬂat lines around $10^{-30}$ are the consequence of the machine precision of the computer which was used. The inverse temperatures $\beta=7.9$ and $\beta =29.4$ yield the line of dynamical instability $\mu_\gamma=1$ for the random matrix models (i) and (ii), respectively.}\label{fig2}
\end{figure}

In Figure~\ref{fig1}
we illustrate the discrepancy between theory and simulations for the two-time covariances $\mathcal C_{\gamma_k}(t,s)$ with respect to the two-time relative-squared-error 
\begin{equation}
\delta \mathcal C_{\gamma_k}(t,s)\doteq\left(\frac{\mathcal C_{\gamma_k}(t,s)-\frac{1}{N_k}\matr \gamma_k(t)^\top \matr \gamma_k(s)}{\mathcal C_{\gamma_k}(t,s)}\right)^2.
\end{equation}
For illustration, the necessary order parameters for the random matrix model (i) are computed by the algorithm considerations described in Section~\ref{comp_order}. Figure~\ref{fig2} illustrates the analytical convergence rate of the algorithm. 

Since we assume that the large-system limit $N_1,N_2\to\infty$ is taken before the long-time limit
$t \to \infty$, we typically get excellent agreement between theoretical predictions and simulations on single instances for
finite-time properties of large systems. However, as the model parameters approach  the dynamical instability $\mu_\gamma=1$, the discrepancy between theory and simulations increases for large times. For example, in Figure~\ref{fig2} we can see that close to the instability the analytical result does not provide accurate results for the random matrix model (i). On the other hand, for the random matrix model (ii) the analytical results give a better approximation. This can be explained by the fact that the system shows smaller fluctuations given that the empirical spectral distribution of the random matrix model (ii) is non-random.

\section{Summary and Outlook}
In this paper, we have introduced and analyzed a new message-passing algorithm for computing
the magnetisations of an RBM Ising network with random coupling matrices. We have assumed
that couplings are drawn at random from a bi-rotation invariant statistical ensemble.
The motivation to study the model with this fairly complex family of ensembles
is the fact that couplings which are learned from applications of RBM in data modeling are expected to inherit
statistical dependencies from the data.We have derived TAP mean-field equations for the magnetisations  for this
class of RBMs and computed static order parameters of the model using the replica
method. We developed a new message-passing algorithm for an iterative
computation of the magnetisations and analyzed its performance in the large system limit.
The algorithm becomes efficient by the fact that a necessary order parameter
can be precomputed from the replica solution.To overcome the problem of performing the quenched averages over the couplings
in the bipartite graph of the model we applied the technique of ``rectangular spherical integration''.
We have shown that the algorithm is globally convergent from certain random initial conditions
as long as a specific criterion which  coincides with the stability of the TAP equations is fulfilled.
We also computed analytical results for the rate of convergence.

We have restricted ourselves to the theoretical analysis of the RBM with a fixed
ensemble of random couplings. It remains to be shown
by future work if the assumption of bi-rotation invariant random matrices, which neglects the 
effect of ``interesting'', non-random eigenvectors, is robust enough to be applicable to RBM
training on real data. An interesting, but more challenging problem would be a
complete theoretical study of RBM training where couplings are developing over time
as learning by gradient descent proceeds.

\section*{Acknowledgment}
This work was funded by the German Research Foundation, Deutsche Forschungsgemeinschaft (DFG), under Grant No. OP 45/9-1. 

\appendix

\section{Useful expressions involving the R-transform}\label{Apre}
We will relate the derivatives $I'(x)$ and $I''(x)$ with the R-transform of free probability \cite{Hiai}. These relations will be useful for deriving the representation of the fixed-point equations of the necessary order parameters in terms of the Green-function, see Section~\ref{comp_order}. Moreover, we will use these relations to state certain random-matrix results involving the R-transforms in terms of $I'(x)$ and/or $I''(x)$. 

The R-transform of the limiting spectral distribution of $\matr W\matr W^\top$ is defined by \cite{Hiai}
\begin{equation}
{\rm R}_{\matr W}(\omega)={\rm G}^{-1}_\matr W(\omega)-\frac{1}{\omega} \label{rtrs}
\end{equation}
where ${\rm G}^{-1}_\matr W$ is the inverse (w.r.t. functional decomposition) of the Green function ${\rm G}_\matr W$ \eqref{Gf}. Furthermore, from \eqref{rtrs} we have the derivative 
\begin{equation}
{\rm R}_{\matr W}'(\omega)=\frac{1}{{\rm G}_{\matr W}'({\rm G}^{-1}_\matr W(\omega))}+\frac{1}{\omega^2}.\label{drtrs}
\end{equation}

As $\psi_1$ and $\psi_2$ are stationary in \eqref{Ix} we have the identities
\begin{align}
\frac{1}{\psi_2}&=\int \frac{{\rm dP}_{\matr W}(t)}{\psi_1\psi_2-xt}\label{good1}\\
\psi_1 &= \alpha\psi_2+(1-\alpha)\label{good2}.
\end{align}
From \eqref{good1} and \eqref{good2} we then obtain respectively
\begin{align}
\psi_1(x)&=\frac{x}{\psi_2(x)}{\rm R}_{\matr W}(\frac{x}{\psi_2(x)})+1\label{psi1}\\
\psi_2(x)&=\frac{x}{\psi_1(x)}{\rm R}_{\matr W^\top}(\frac{x}{\psi_1(x)})+1\label{see3}.
\end{align}
Moreover, we have a formula for the derivative ${I}'(x)$ in terms of the R-transform as 
\begin{equation}
{I}'(x)=\frac{1}{\psi_2(x)}{\rm R}_{\matr W}(\frac{x}{\psi_2(x)}).\label{frtrs}
\end{equation}

By using \eqref{psi1}, \eqref{see3} and \eqref{frtrs} we write the derivatives
\begin{align}
\psi'_1(x)&
=\left[I'(x)+\frac{x}{\psi_2(x)^2}{\rm R}'_{\matr W}(\frac{x}{\psi_2(x)})\right]\left(1- \frac{x\psi_2'(x)}{\psi_2(x)}\right)\label{see1}\\
\psi'_2(x)&=\left[\frac{I'(x)}{\alpha}+\frac{x}{\psi_1(x)^2}{\rm R}'_{\matr W^\top}(\frac{x}{\psi_1(x)})\right]\left(1- \frac{x\psi_1'(x)}{\psi_1(x)}\right)\label{see2}.
\end{align}
Moreover, from \eqref{good2}, \eqref{psi1} and \eqref{frtrs} we point out the identities
\begin{align}
\psi_1'(x)=\alpha\psi_2'(x)={I}'(x)+x{I}''(x)
\end{align}
Using these results we have the expressions
\begin{align}
\frac{x}{\psi_2(x)^2}{\rm R}'_{\matr W}(\frac{x}{\psi_2(x)})&=  \frac{\alpha\psi_2(x)(I'(x)+\chi I''(x))}{\alpha-x^2 I''(x)}-I'(x)\label{u1}\\
\frac{x}{\psi_1(x)^2}{\rm R}'_{\matr W^\top}(\frac{x}{\psi_1(x)})&=\frac{\psi_1(x)(I'(x)+\chi I''(x))}{\alpha-\alpha x^2 I''(x)}-\frac{I'(x)}{\alpha}.\label{u2}
\end{align}
\section{The replica-symmetry calculation of the free energy}\label{rep}
For an integer $p$, we will first compute 
\begin{equation}
F(p)\doteq\lim_{N\to \infty}\frac 1 N \ln \mathbb E[Z^p].
\end{equation}
Specifically, performing the \emph{rectangular-spherical integration} method and the saddle point method  one can show that
\begin{align}
F(p)=& \operatorname*{extr}_{\{\mathcal Q_k,\mathcal {\hat Q}_k\}}\left\{\ln Z_1(\mathcal {\hat Q}_1)+\alpha\ln Z_2(\mathcal {\hat Q}_2)+\frac 1 2 {\rm tr}(I(\mathcal Q_1\mathcal Q_2))
\nonumber\right. \\&\left.\hspace{2cm}-\sum_{a<b} \mathcal {\hat Q}_1(a,b)\mathcal Q(a,b)-\alpha\sum_{a<b} \mathcal {\hat Q}_2(a,b)\mathcal Q_2(a,b)\right\}. \label{Fp}
\end{align}
Here, we have introduced the partition functions (for $k=1,2$)
\begin{equation}
Z_k(\mathcal {\hat Q}_k)\doteq \sum_{\{s(a)=\mp 1\}} {\mathrm e}^{\sum_{a<b}\mathcal {\hat Q}_k(a,b)s(a)s(b)+h_k\sum_{a}s(a)}.
\end{equation}
Moreover, $\mathcal Q_k$ and $\mathcal {\hat Q}_k$ are all $p \times p$ matrices which satisfy the equalities 
\begin{align}
\mathcal Q_k(a,b)&=\mathbb E[s(a)s(b)]_{\mathcal Z_k(\mathcal Q_k)}\\
\mathcal {\hat Q}_1&=I'(\mathcal Q_2\mathcal Q_1)\mathcal Q_2\\
\mathcal {\hat Q}_2&=\frac {I'(\mathcal Q_1\mathcal Q_2)\mathcal Q_1}{\alpha }.\label{hatrep}
\end{align} 
We now assume the replica symmetries 
\begin{equation}
Q_k(a,b)=q_k,\forall a\neq b.
\end{equation}
These imply that $\hat Q_k(a,b)=\hat q_k,\forall a\neq b$. Thereby, $F(p)$ in \eqref{Fp} reads as
\begin{align}
&\operatorname*{extr}_{\{q_k,\hat q_k\}}\left\{\ln \mathbb E[(2\cosh(h_1+\sqrt{\hat q_1}u))^p]+\alpha \ln \mathbb E[(2\cosh(h_2+\sqrt{\hat q_2}u))^p]-\frac{p(p-1)}{2}(q_1\hat q_1+\alpha q_2\hat q_2)+\right. \nonumber \\ &\left. \qquad + \frac{1}{2}I((p-1)^2q_1q_2+(p-1)(q_1+q_2)+1)+\frac{p-1}{2}I(q_1q_2-(q_1+q_2)+1)\right\}.
\end{align}
Then, we obtain the RS approximation of the free energy as
\begin{align}
F'(0)&= \operatorname*{extr}_{\{\chi_k,\hat q_k\}}\mathbb E[\ln 2\cosh(h_1+\sqrt{\hat q_1}u)]+\alpha  \mathbb E[\ln 2\cosh(h_2+\sqrt{\hat q_2}u)]+\nonumber \\
&\qquad-\frac{1}{2}(\chi_1\hat q_1+\alpha\chi_2{\hat q_2}) +\frac 1 2 I(\chi_1\chi_2)-(\chi_1+\chi_2+2\chi_1\chi_2)I'(\chi_1\chi_2)\label{rsf}
\end{align}
where for convenience we consider substitutions $\chi_k\doteq 1-q_k$. Furthermore, extremisations of \eqref{rsf}  w.r.t.$\{\chi_k,\hat q_k\}$ yield the fixed-point equations
\begin{subequations}
	\begin{align}
	\chi_1&=\mathbb E[\tanh'(h_1+ \sqrt{\hat q_1}u)]\\
	\chi_2&=\mathbb E[\tanh'(h_2+\sqrt{\hat q_2}u)]\\
	\hat q_1&=\chi_2^2(1-\chi_1){I}''(\chi)+(1-\chi_2)(I'(\chi)+\chi I''(\chi))\\
	\hat q_2&=\frac {\chi_1^2(1-\chi_2){I}''(\chi)+(1-\chi_1)(I'(\chi)+\chi I''(\chi))}{\alpha}.
	\end{align}
\end{subequations} 


\section{Derivation of the TAP equations}\label{dervtap}
We manipulate the exact marginalized distributions of the pairs $(s_{1i},s_{2j})$ as 
\begin{align}
p(s_{1i},s_{2j}\vert \matr W,h_1,h_2)&=\frac 1 Z {\mathrm e}^{s_{1i}h_{1}+s_{2j}h_{2}}\int {\rm d}\matr s_{\backslash i,j}\;{\mathrm e}^{s_{1i}(\sum_{k\neq {j}}W_{ik}s_{2k})+s_{2j}(\sum_{k\neq i}W_{kj}s_{1k})}\nonumber \\
&\quad\qquad \qquad \qquad\qquad\times \underbrace{{\mathrm e}^{\matr s_{1\backslash i}^\top\matr W_{\backslash ij}\matr s_{2\backslash j}+\matr h_{1\backslash i}^\top \matr s_{1\backslash i}+\matr h_{2\backslash j}^\top \matr s_{2\backslash j}}}_{\doteq Z_{\backslash i,j}p(\matr s_ {\backslash i,j}\vert \matr W_{\backslash ij},h_{1},h_{2})}\\
&= \frac {Z_{\backslash i,j}} Z {\mathrm e}^{s_{1i}h_1+s_{2j}h_2}\int {\rm d}\theta_i {\rm d}\theta_j\;{\mathrm e}^{s_{1i}\theta_i+s_{2j}\theta_j}p(\theta_i,\theta_j)
\end{align}
where we have defined the \emph{bi-variate cavity distributions} 
\begin{equation}
p(\theta_i,\theta_j)\doteq \int {\rm d}\matr s_{\backslash i,j}\;\delta (\theta_i-\sum_{k\neq j}W_{ik}s_{2k})\delta(\theta_j-\sum_{k\neq i}W_{kj}s_{1k})p(\matr s_ {\backslash i,j}\vert \matr W_{\backslash ij},  h_{1}, h_{2}).
\end{equation}
Following arguments of \cite[Chapter V.3]{Mezard} we assume weak dependencies between the spins variables
expressed by the block covariance matrix
\begin{equation}
\left[\begin{array}{cc}
\matr \chi_{11}&\matr\chi_{12}\\
\matr \chi_{21}&\matr \chi_{22}
\end{array}\right]_{nn'}= O(1/\sqrt{N_1}),\quad \forall n\neq n'  \label{weak} \quad 
\end{equation}
where we have defined the covariance matrices
\begin{equation}
\matr \chi_{kk'}= \mathbb E [\matr s_{k}\matr s_{k'}^\top]-\matr m_{k}\matr m_{k'}^\top, \qquad k,k'\in\{1,2\}.
\end{equation}
We then approximate the cavity distributions as 
\begin{equation}
p(\theta_i,\theta_j)\approx \mathcal N(\theta_i\vert \gamma_{1i}, v_{1i})\mathcal N(\theta_j\vert \gamma_{2j}, v_{2j})\quad \forall i,j.\label{cavapprox}
\end{equation}
Later in \ref{derstab}, we will sort out a stability analysis for the weak-dependency assumption \eqref{weak}.
 The Gaussian approximations \eqref{cavapprox} lead immediately to 
\begin{subequations}
\label{adatap}
\begin{align}
\matr m_1&=\tanh(h_1+\matr \gamma_1)\\
\matr m_2&=\tanh(h_2+\matr \gamma_2)\\
\matr \gamma_1&=\matr W\matr m_2-\matr V_1\matr m_1\\
\matr \gamma_2&=\matr W^\top \matr m_1-\matr V_2\matr m_2
\end{align}	
\end{subequations}
where for convenience we have introduced the diagonal matrices of the cavity variances  $\matr V_k\doteq {\rm diag}(v_{k1},\cdots,v_{kN_k})$. Moreover, by linear-response the approximations \eqref{cavapprox}  yield
\begin{equation}
\left[\begin{array}{cc}
\matr \chi_{11}&\matr\chi_{12}\\
\matr \chi_{21}&\matr \chi_{22}
\end{array}\right]= \left[\begin{array}{cc}
(\matr \Lambda_1-\matr W\matr \Lambda_2^{-1}\matr W^\top)^{-1}&(\matr \Lambda_1-\matr W\matr \Lambda_2^{-1}\matr W^\top)^{-1}\matr W\matr \Lambda_2^{-1}\\
\matr \Lambda_2^{-1}\matr W^\top(\matr \Lambda_1-\matr W\matr \Lambda_2^{-1}\matr W^\top)^{-1}&(\matr \Lambda_2-\matr W^\top\matr \Lambda_1^{-1}\matr W)^{-1}
\end{array}\right]\label{matx-cov}
\end{equation}
where we have introduced  the diagonal matrices $\matr \Lambda_k$ with the diagonal entries
\begin{equation}
(\matr \Lambda_k)_{nn}=\frac{1}{(\matr\chi_{kk})_{nn}}+{({\matr V}_{k})}_{nn}=\frac{1}{\tanh'(h_{k}+\gamma_{kn})}+{({\matr V}_{k})}_{nn} \quad \forall n.\label{cavmat}
\end{equation}
The equations  \eqref{adatap}--\eqref{cavmat} form together the so-called \emph{adaptative TAP equations} \cite{Adatap} for the spin-glass model \eqref{Gibbs2}. 
\subsection{Self-averaging property of the cavity variances} 
We will use the concept of asymptotic freeness of random matrices to show that the cavity variances are asymptotically self-averaging. Specifically, under certain technical assumptions we may assume that  the bi-rotation invariant $\matr W$ is asymptotically free of the diagonals $\{\matr\Lambda_1,\matr \Lambda_2\}$ \cite{Hiai}. Doing so will lead to 
\begin{equation}
{\matr V}_{1}\simeq \chi_2I'(\chi){\bf I} \quad \text{and} \quad {\matr V}_{2}\simeq  \frac{\chi_1I'(\chi)}{\alpha}{\bf I} \label{selfcav}.
\end{equation}
where we have defined
\begin{equation}
\chi_k\doteq\lim_{N_{k}\to \infty} \mathbb E_\matr W[\tanh'(h_{k}+\gamma_{kn})]
\end{equation}
Note, that plugging \eqref{selfcav} into the \emph{adaptative TAP equations} \eqref{adatap} yields the TAP equations \eqref{tap2} given that $\{\chi_k\}$ are the solutions of \eqref{s.order}. The RS calculation of $\{\chi_k\}$ \eqref{s.order} can be independently read off from the results of the DF analysis.

\begin{remark}\label{remarkfree}
Let us introduce the variables
\begin{align}
d_{1n}&\doteq((\matr \Lambda_1-\matr W\matr \Lambda_2^{-1}\matr W^\top)^{-1})_{nn}-((\matr \Lambda_1)_{nn}- \chi_2I'(\chi))^{-1}\quad \forall n.\\
d_{2n}&\doteq((\matr \Lambda_2-\matr W^\top\matr \Lambda_1^{-1}\matr W)^{-1})_{nn}-((\matr \Lambda_2)_{nn}-\frac{\chi_1I'(\chi)}{\alpha})^{-1}\quad \forall n.
\end{align}	
Furthermore, let the matrices $\matr W\matr W^\top$, $\matr\Lambda_1$ and $\matr \Lambda_2$ have a limiting spectral distribution,~each. Moreover, let the bi-rotation invariant random matrix $\matr W$ be 
	asymptotically free of the diagonals $\{\matr\Lambda_1,\matr \Lambda_2\}$. Then, we have 
	\begin{equation}
	\lim_{N_k\to \infty}\mathbb E[d_{kn}^2]=0 \quad \forall k,n \label{cavself}
	\end{equation}
where the expectation is taken over random matrix $\matr W$.
\end{remark}
Note that \eqref{cavself} implies \eqref{selfcav} in a $L^2$ norm sense. For an explicit derivation of the Remark~\ref{remarkfree} we refer to the derivation of \cite[Theorem~1]{CakmakOpper18}.	

\subsection{The stability of the TAP Equations}\label{derstab}
The spin-covariance matrix of the TAP equations \eqref{tap2}  
coincides with the matrix \eqref{matx-cov} such that the diagonal elements of the (diagonal) matrices $\matr \Lambda_k$ are substituted by
\begin{align}
(\matr \Lambda_1)_{ii}&=\frac{1}{\tanh'(h_{1}+\gamma_{1i})}+\chi_2I'(\chi)\\
(\matr \Lambda_2)_{jj}&=\frac{1}{\tanh'(h_{2}+\gamma_{2j})}+\frac{\chi_1I'(\chi)}{\alpha}.
\end{align}
Our goal is to derive the stability criterion for the condition 
\begin{equation}
\lim_{N_1\to \infty}N_1\mathbb E[(\matr \chi_{kk'})_{nn'}^2]=O(1) \quad  \forall n\neq n',k,k'\label{show}
\end{equation}
where the expectation is taken over random matrix $\matr W$. By symmetry we have 
\begin{align}
\frac 1{N_k}\mathbb E[{\rm tr}(\matr \chi_{kk}\matr \chi_{kk})]&=(N_k-1)\mathbb E[(\matr \chi_{kk})_{nn'}^2]+\mathbb E[(\matr \chi_{kk})_{nn}^2]~~  n\neq n'.\\
\frac 1{N_1}\mathbb E[{\rm tr}(\matr \chi_{12}\matr \chi_{21})]&=N_2\mathbb E[(\matr \chi_{12})_{ij}^2].
\end{align}
Here,  
Hence, \eqref{show} holds if and only if we have
\begin{align}
\chi_{kk}^{(2)}\doteq \lim_{N_k\to \infty}\frac 1{N_k}\mathbb E[{\rm tr}(\matr \chi_{kk}\matr \chi_{kk})]=O(1)\\
\chi_{12}^{(2)}\doteq\lim_{N_1\to \infty} \frac 1{N_1}\mathbb E[{\rm tr}(\matr \chi_{12}\matr \chi_{21})]=O(1).
\end{align} 
\begin{remark}\label{remat}
Let the matrices $\matr W\matr W^\top$, $\matr\Lambda_1$ and $\matr \Lambda_2$ have a limiting spectral distribution, each. Furthermore, let  $\matr W$ be asymptotically free of the diagonals $\{\matr\Lambda_1,\matr \Lambda_2\}$. Then, we have
\begin{align}
\chi_{kk}^{(2)}&=\frac{\eta_k}{1-\eta_k{\rm R}'_k}\\
\chi_{12}^{(2)}&=
\frac{\psi_1'[\psi_1\psi_2-\chi(\psi_2\psi_1'+\psi_1\psi_2')]}{[\psi_1-\chi\psi_1'][\psi_2-\chi\psi_2']}\chi_{11}^{(2)}\chi_{22}^{(2)}.
\end{align}
Here, $\chi=\chi_1\chi_2$ and $\{{\rm R}_k'\}$ \& $\{{\psi}_k\}$ are defined as in \eqref{ATr} \& \eqref{psi}, respectively. Furthermore, we have $\psi_1'\doteq I'(\chi)+\chi I''(\chi)$ and $\psi_2'\doteq \frac{1}{\alpha}\psi_1'$. Moreover, we have defined 
\begin{align}
\chi_k\doteq &\lim_{N_k\to \infty}\mathbb E[\tanh'(h_{k}+\gamma_{kn})]\\
\eta_k\doteq &\lim_{N_k\to \infty}\mathbb E[(\tanh'(h_{k}+\gamma_{kn})^2].
\end{align}
\end{remark}
Remark~\ref{remat} implies that \eqref{show} holds if and only if 
\begin{equation}
\eta_k{\rm R}_k'<1, \quad k=1,2
\end{equation}
given that the critical cases $\{\psi_k=\chi\psi_k'\}$ (for $\chi_{12}^{(2)}$) are fulfilled
as $\{{\rm R}_k'\}$ tend to infinity.

We next present a sketch of the derivation of Remark~\ref{remat}. To this end, we introduce  generating functions
\begin{align}
\chi_{kk}(\omega)&\doteq\lim_{N_k\to \infty}\frac 1{N_k}\mathbb E[{\rm tr}\left((\matr \Lambda_k-\omega{\bf I}-\matr W\matr \Lambda_{k'}^{-1}\matr W^\top)^{-1}\right)] \quad k\neq k'\label{gf1}\\
\chi_{12}(\omega)&\doteq \lim_{N_1\to \infty} \frac 1{N_1}\mathbb E[{\rm tr}\left((\matr \Lambda_1-\matr W(\matr \Lambda_2-\omega{\bf I})^{-1}\matr W^\top)^{-1}\right)]\label{gf2}.
\end{align}
In particular, it is easy to show that
\begin{subequations}
\label{iden2}
\begin{align}
\chi_{kk}^{(2)}&=\chi_{kk}'(0)\\\chi_{12}^{(2)}&=\chi_{12}'(0)
\end{align}	
\end{subequations}
where e.g. $\chi_{11}'$ stands for the derivative of $\chi_{11}$. Hence, we can first simplify the generating functions \eqref{gf1} and \eqref{gf2} using the asymptotic freeness assumption and then invoke the identities \eqref{iden2}. We skip the explicit and lengthy calculation. Instead, we refer the reader to the arguments of \cite[Remark~1]{CakmakOpper18} and \cite[Eq. (C.39)-(C.42)]{cakmak2016random}. These references refer to the random matrix results in terms of the R-transform. Using the R-transform relations in \ref{Apre}, they can be reformulated in terms of $I'(\chi)$ and/or $I''(\chi)$.


\section{Derivations of the results on the DF analysis}\label{derDF}

We will first re-express the moment-generating functional \eqref{gff} in such way that the disorder average can be conveniently performed using the method of \emph{rectangular spherical integration}. To this end, for the sake of compactness of notations, we introduce the scalars
\begin{equation}
\lambda_k\doteq  \frac{\psi_k}{\chi_k}, \quad k=1,2.
\end{equation}
Furthermore, we introduce the (fixed) matrices
\begin{align}
\matr \Lambda \doteq \left(\begin{array}{cc} \lambda_1{\bf I} & \matr 0\\
\matr 0&\lambda_2{\bf I}
\end{array}\right),~~~~\matr D \doteq \left(\begin{array}{cc}
\chi_1{\bf I} & \matr 0\\
\matr 0&\chi_2{\bf I}
\end{array}\right)~~\text{and}~~\matr J \doteq \left(\begin{array}{cc}
\matr 0 & \matr W\\
\matr W^\top&\matr 0
\end{array}\right).
\end{align}
Hence, we can write
\begin{equation}
\matr A=\matr X^{-1}-{\bf I} ~~\text{with}~~  \matr X\doteq(\matr \Lambda-\matr J)\matr D.
\end{equation}
Then, by using the property of Dirac-delta function $\delta(\matr y)=\vert \matr X\vert \delta(\matr X\matr y)$ we have 
\begin{align}
\delta[\matr \gamma(t)-\matr A\tilde{\matr \gamma }(t) ]&= \vert \matr X\vert \delta[\tilde{\matr \gamma}(t)-\matr X(\matr \gamma(t)+\tilde {\matr \gamma}(t))]~~\text{with}~~ \tilde{\matr\gamma}(t)\doteq f(\matr \gamma(t-1))\label{t1}.
\end{align}
By invoking respectively \eqref{t1} and the Dirac-delta function in terms of its characteristic function we write
\begin{align}
Z_{ij}(\{l_{1}(t),l_{2}(t)\})&=\int \prod_{t=1}^{T}{\rm d}\matr m(t){\rm d}\tilde{\matr\gamma}(t){\rm d}\matr\gamma(t)\;
\delta [\tilde {\matr \gamma}(t)- f({\matr \gamma}(t-1))]\delta[\matr m (t)-\matr D(\matr \gamma(t)+\tilde {\matr \gamma}(t))]\nonumber\\& \hspace{0.1cm}\qquad \qquad \times\vert\matr X\vert\delta\left[\tilde{\matr \gamma}(t)- (\matr \Lambda-\matr J)\matr m(t)\right] {\mathrm e}^{{\rm i}\sum_{t=0}^{T}[l_{1}(t)\gamma_{1i}(t)+l_{2}(t)\gamma_{2j}(t)]} \\
=&c\int \prod_{t=1}^{T}{\rm d}\hat{\matr \gamma}(t){\rm d}\matr m(t){\rm d}\tilde{\matr\gamma}(t){\rm d}\matr\gamma(t)\;
\delta[\tilde{\matr \gamma}(t)- f({\matr \gamma}(t))]\delta[\matr m (t)-\matr D(\matr \gamma(t)+\tilde{\matr \gamma}(t))]\nonumber\\&\hspace{1.4cm}\qquad \qquad\times {\mathrm e}^{{\rm i}\hat{\matr \gamma}(t)^\top\left[\tilde{\matr \gamma}(t)- (\matr \Lambda-\matr J)\matr m(t)\right]} {\mathrm e}^{{\rm i}\sum_{t=0}^{T}[l_{1}(t)\gamma_{1i}(t)+l_{2}(t)\gamma_{2j}(t)]}
\end{align}
where the determinant $\vert\matr X\vert$ does not depend on $\matr O$ and $\matr V$ and $c$ stands for a \emph{constant term} for ensuring the normalization property $Z_{ij}(\{0,0\})=1$.
\subsection{Disorder average}
Consider the decompositions  $\matr m(t)\doteq{\small \left[\begin{array}{c}
	\matr m_1(t)\\
	\matr m_2(t)
	\end{array}\right]}$ and $\hat{\matr \gamma}(t)\doteq{\small \left[\begin{array}{c}
\hat{\matr \gamma}_1(t)\\
\hat{\matr \gamma}_2(t)
	\end{array}\right]}$ where the vectors $\matr m_1(t)$ and $\hat {\matr \gamma}_1(t)$ are of dimensions $N_1\times 1$. Furthermore, we introduce the $N_1 \times T$ matrices ${\matr {X}}_1$  and $\matr {\hat X}_1$ and the $N_2\times T$ matrices ${\matr {X}}_2$  and $\matr {\hat X}_2$ with the entries 
\begin{align}
(\matr X_{k})_{it}\doteq\frac{m_{ki}(t)}{\sqrt{N_k}}~~\text{and}~~(\hat{\matr X}_{k})_{it}\doteq \frac{{\rm i}\hat \gamma_{ki}(t)}{\sqrt{N_k}}.
\end{align}
So that, we write
\begin{equation}
{\mathrm e}^{{\rm i}\sum_{t\leq T}\hat{\matr \gamma}(t)^\top \matr J\matr m(t)}={\mathrm e}^{\sqrt{N_1N_2}{\rm tr}(\matr W\matr Q)} ~~\text{with}~~ {\matr Q}\doteq {\matr X}_2\hat{\matr X}_1^\top+\hat{\matr X}_2\matr X_1^\top.
\end{equation}

We express the generating function $I(x)$ in \eqref{Ix} in terms of a formal power series as 
\begin{equation}
I(x)=\sum_{n=1}^{\infty}\frac{c_n}{n}x^n.\label{powerI}
\end{equation}
given that $I(0)=0$. Then, we have 
\begin{equation}
\mathbb E\left[{\mathrm e}^{\sqrt{N_1N_2}{\rm tr}(\matr W\matr Q)}\right]_{\matr O, \matr V}= {\mathrm e}^{\frac{N_1}{2}\left(\epsilon_{N_1}+\sum_{n=1}^{\infty}\frac{c_n}{n}{\rm tr}((\matr Q\matr Q^\top)^n)\right)}
\end{equation}
with the constant term $\epsilon_{N_1}\to0$ as $N_1\to \infty$. We will evaluate ${\rm tr}((\matr Q\matr Q^\top)^n)$ in terms of the $T\times T$ order parameter matrices 
\begin{align}
\mathcal {G}_k&\doteq\matr{X}_k^\top \matr {\hat X}_k \\
\mathcal{C}_k&\doteq\matr{X}_k^\top \matr {X}_k \\
\mathcal{\tilde{C}}_k&\doteq\matr{\hat X}_k^\top \matr {\hat X}_k.
\end{align}
Specifically, we have
\begin{align}
f_n(\mathcal {G}_1,\mathcal {G}_2,\mathcal{C}_1,\mathcal{C}_2,\mathcal{\tilde{C}}_1, \mathcal{\tilde{C}}_2)\doteq{\rm tr}((\matr Q\matr Q^\top)^n)= {\rm tr}\left\{\left[\left(\begin{array}{cc}
\mathcal{\tilde{C}}_1 & \mathcal {G}_1^\top\\
\mathcal {G}_1&\mathcal{C}_1
\end{array}\right)\left(\begin{array}{cc}
\mathcal{{C}}_2 & \mathcal {G}_2\\
\mathcal {G}_2^\top&\mathcal{\tilde C}_2
\end{array}\right) \right]^n\right\}.
\end{align}

We will be interested in calculating the trace of the power of the matrix
\begin{equation*}
\left(\begin{array}{cc}
\mathcal{\tilde{C}}_1 & \mathcal {G}_1^\top\\
\mathcal {G}_1&\mathcal{C}_1
\end{array}\right)\left(\begin{array}{cc}
\mathcal{{C}}_2 & \mathcal {G}_2\\
\mathcal {G}_2^\top&\mathcal{\tilde C}_2
\end{array}\right)=\left(\begin{array}{cc}
\underbrace{\mathcal{\tilde{C}}_1\mathcal C_2+\mathcal G_1^\top \mathcal G_2^\top}_{\mathcal A} & \underbrace{\mathcal{\tilde{C}}_1\mathcal G_2+\mathcal G_1^\top \tilde{\mathcal C}_2}_{\mathcal B}\\
\underbrace{\mathcal {G}_1\mathcal C_2+\mathcal C_1\mathcal G_2^\top}_{\mathcal C}&\underbrace{\mathcal{G}_1\mathcal G_2+\mathcal C_1\tilde{\mathcal C}_2}_{\mathcal D}
\end{array}\right)
\end{equation*}
at the saddle-point values $\tilde{\mathcal C}_1=\matr 0$ and  $\tilde{\mathcal C}_2=\matr 0$. In particular, we have
\begin{equation}
{\rm tr}\left[\left(\begin{array}{cc}
\mathcal{A} & \mathcal {B}\\
\mathcal {C}&\mathcal{D}
\end{array}\right)^n\right]={\rm tr}\left(\mathcal A^n+\mathcal D^n+n\mathcal B\sum_{k=1}^{n-1}\mathcal D^{k-1}\mathcal C\mathcal A^{n-1-k}\right)+\mathcal{SP}(\mathcal A,\mathcal B,\mathcal C,\mathcal D)\label{remnice}
\end{equation}
where 
\begin{equation}
\left.\frac{\partial \mathcal{SP}(\mathcal A,\mathcal B,\mathcal C,\mathcal D)}{\partial\mathcal{B}}\right\vert_{\mathcal {B}=\matr 0}=\matr 0.
\end{equation}
In other words, at the saddle-point values $\mathcal {B}=\matr 0$ and the term ${\mathcal {SP}}(\mathcal A,\mathcal B,\mathcal C,\mathcal D)$ does not contribute to saddle--point equations.
\subsection{Saddle-point analysis}
We introduce the \emph{single-site} generating functional
\begin{align}
Z_{1}(\{l(t)\},\mathcal {\hat G}_1,\mathcal {\hat C}_1,\mathcal {\hat {\tilde C}}_1)\doteq & c \int {\rm d}\gamma_1(0)\;\mathcal N(\gamma_1(0)\vert 0,\hat q_1) 
\prod_{t=1}^{T}{\rm d}\tilde\gamma_1(t) {\rm d}\gamma_1(t){\rm d}m_1(t){\rm d}{\hat\gamma}_1(t)\;\delta[\tilde\gamma_1(t)- f_1({\gamma}_1(t-1))]   \nonumber \\
&
\quad \times  \delta[m_1(t)-\chi_1 (\gamma_1(t)+\tilde{\gamma}_1(t))]
{\mathrm e}^{ {\rm i}\hat\gamma_1(t)(\tilde{\gamma}_1(t)-\lambda_1m_1(t))}\nonumber \\
&\quad \times {\mathrm e}^{-\sum_{(t,s)}[-{\rm i}\mathcal{\hat G}_1(t,s)m_1(t){\hat\gamma}_1(s)+ {\rm i}\mathcal {\hat C}_1(t,s)m_1(t){m}_1(s)+\mathcal{\hat{\tilde C}}_1(t,s){\hat\gamma}_1(t) {\hat\gamma}_1(s)]}\nonumber \\
&\quad  \times {\mathrm e}^{ {\rm i}\sum_{t\leq 0}\gamma_1(t)l(t)}.
\end{align}
Here, for example $\mathcal {\hat G}_1(t,s)$ stands for the $(t,s)$ indexed entry of $\mathcal {\hat G}_1$. Similarly, we define $Z_{2}(\{l(t)\},\mathcal {\hat G}_2,\mathcal {\hat C}_2,\mathcal {\hat {\tilde C}}_2)$. Thereby, we can write the averaged generating functional in the form
\begin{align}
\mathbb E[Z_{ij}(\{l_1(t),l_2(t)\})]= & c\int {\rm d} \mathcal G_1 {\rm d}\mathcal {\hat G}_1 {\rm d} \mathcal C_1 {\rm d}\mathcal {\hat C}_1 {\rm d} \mathcal {\tilde C}_1 {\rm d}\mathcal{\hat {\tilde C}}_1  {\rm d} \mathcal G_2 {\rm d}\mathcal {\hat G}_2 {\rm d} \mathcal C_2 {\rm d}\mathcal {\hat C}_2 {\rm d} \mathcal {\tilde C}_2 {\rm d}\mathcal{\hat {\tilde C}}_2 \;\nonumber \\
& \qquad \times Z_{1}(\{l_1(t)\},\mathcal {\hat G}_1,\mathcal {\hat C}_1,\mathcal {\hat {\tilde C}}_1)Z_{2}(\{l_2(t)\},\mathcal {\hat G}_2,\mathcal {\hat C}_2,\mathcal {\hat {\tilde C}}_2)\nonumber \\& \qquad \times {\mathrm e}^{\frac{N_1}{2}\left(\epsilon_{N_1}+\sum_{n\geq 1} \frac{c_n}{n} f_n(\mathcal {G}_1,\mathcal {G}_2,\mathcal{C}_1,\mathcal{C}_2,\mathcal{\tilde{C}}_1, \mathcal{\tilde{C}}_2)\right)} \nonumber \\
&\qquad\times   {\mathrm e}^{N_1\sum_{(t,s)}[-\mathcal{\hat G}_1(t,s) \mathcal G_1(t,s)+{\rm i}\mathcal{\hat C}_1(t,s)\mathcal C_1(t,s) -\mathcal{\hat{\tilde C}}_1(t,s) \mathcal{\tilde C}_1(t,s)] }\nonumber\\
&\qquad \times  {\mathrm e}^{N_2\sum_{(t,s)}[-\mathcal{\hat G}_2(t,s) \mathcal G_2(t,s)+{\rm i}\mathcal{\hat C}_2(t,s)\mathcal C_2(t,s) -\mathcal{\hat{\tilde C}}_2(t,s) \mathcal{\tilde C}_2(t,s)]}.
\end{align}
In the large system limit, we can perform the integration over $\{\mathcal G_k, \mathcal {\hat G}_k,\mathcal C_k, \mathcal {\hat C}_k,\mathcal {\tilde C}_k, \mathcal {\hat {\tilde C}}_k\}$ with the saddle point methods. Doing so yields (for $k=1,2$):
\begin{align}
\mathcal G_k(t,s)&={\rm i}\mathbb E[m_k(t)\hat\gamma_k(s)]_{Z_k}\label{{MG}}\\
\mathcal C_k(t,s)&=\mathbb E[m_k(t)m_k(s)]_{Z_k}\label{mag}\\
\mathcal{\tilde C}_k(t,s)&=-\mathbb E[\hat\gamma_k(t)\hat\gamma_k(s)]_{Z_k} \label{unpsy}
\end{align}
where $\mathbb E[(\cdot)]_{Z_k}$ stands for the expectation with respect to the single-site generating functionals $Z_{k}(\{l(t)\},\mathcal {\hat G}_k,\mathcal {\hat C}_k,\mathcal {\hat {\tilde C}}_k)$.  Furthermore, we consider the solutions $\mathcal{\tilde C}_k=\matr 0$ at the saddle points which yields $\mathcal {\hat C}_k=\matr 0$. Moreover, by invoking \eqref{remnice} we have
\begin{subequations}
	\label{covfs}
	\begin{align}
	2\mathcal {\hat {\tilde C}}_1&=\left[\sum_{n=1}^{\infty} c_n(\mathcal G_2\mathcal G_1)^{n-1}\right]\mathcal C_2+\left[\sum_{n=1}^{\infty}c_n\sum_{k=1}^{n-1}(\mathcal G_1\mathcal G_2)^{k-1}\mathcal C(\mathcal G_1^\top \mathcal G_2^\top)^{n-1-k}\right]\mathcal G_2\\
	2\mathcal {\hat {\tilde C}}_2&=\left[\sum_{n=1}^{\infty}c_n(\mathcal G_1\mathcal G_2)^{n-1}\right]\frac{\mathcal C_1}{\alpha}+\left[\sum_{n=1}^{\infty}c_n\sum_{k=1}^{n-1}(\mathcal G_1\mathcal G_2)^{k-1}\mathcal C(\mathcal G_1^\top \mathcal G_2^\top)^{n-1-k}\right]\frac{\mathcal G_1}{\alpha}.
	\end{align}
\end{subequations}
with noting that $\mathcal C=\mathcal G_1\mathcal C_2+\mathcal C_1\mathcal G_2^\top$. We also get
\begin{subequations}
	\label{ghats}
	\begin{align}
	\mathcal {\hat {G}}_1&=\left[\sum_{n=1}^{\infty} c_n(\mathcal G_2\mathcal G_1)^{n-1}\right]\mathcal G_2=I'(\mathcal G_2\mathcal G_1)\mathcal G_2\\
	\mathcal {\hat {G}}_2&=\left[\sum_{n=1}^{\infty} c_n(\mathcal G_1\mathcal G_2)^{n-1}\right] \frac{\mathcal G_1}{\alpha}=\frac{I'(\mathcal G_1\mathcal G_2)\mathcal G_1}{\alpha}.
	\end{align} 	
\end{subequations}
In these equations, we drop the contributions $\frac{\partial \epsilon_{N_1}}{\partial \mathcal X}$ for $\mathcal  X=\{\mathcal G_k,\mathcal {C}_k,\mathcal {\tilde C}_k,k=1,2\}$ at the saddle point analysis, given that $\epsilon_{N_1}\simeq 0$. 

For convenience, we define $ Z_{k}(\{l_k(t)\})\doteq  Z_{k}(\{l_k(t)\},\mathcal {\hat G}_k,\matr 0,\mathcal {\hat {\tilde C}}_k)$ where the $T\times T$ order matrices  $\{\mathcal {\hat {\tilde C}}_k\}$ and $\{\mathcal {\hat G}_k\}$ are given as in \eqref{covfs} and \eqref{ghats}, respectively. Then, the saddle point analysis leads to 
\begin{equation}
\mathbb E[Z_{ij}(\{l_1(t),l_2(t)\})]\simeq Z_1(\{l_1(t)\})\times  Z_2(\{l_2(t)\}).
\end{equation} 
To integrate the variables $\{\hat \gamma_k(t)\}$, in $Z_{k}(\{l_k(t)\})$ we linearize the quadratic terms in $\hat \gamma_k(t)$ by introducing auxiliary zero-mean Gaussian processes $\{\phi_k(t)\}$  with the $T\times T$ covariance matrices $\mathcal C_{\phi_k}\doteq 2\mathcal {\hat {\tilde C}}_k$, so that we can write
\begin{equation}
{\mathrm e}^{-\sum_{t,s}\mathcal {\hat {\tilde C}}_k(t,s)\hat \gamma_k(t)\hat \gamma_k(s)}=\mathbb E[{\mathrm e}^{-{\rm i}\sum_{t}\hat \gamma_k(t)\phi_k(t)}].
\end{equation}
Then, the single-site generating functionals reads as
\begin{align}
Z_k(\{l(t)\})=\int &{\rm d}\gamma_k(0)\{{\rm d}\phi_k(t)\} \; \mathcal N(\gamma_k(0)\vert 0,\hat q_k)\mathcal N(\phi_k(1),\dots,\phi_k(T)\vert \matr 0, \mathcal C_{\phi_k})\nonumber\\ &\times \prod_{t=1}^{T} {\rm d}\tilde\gamma_k(t) {\rm d}\gamma_k(t){\rm d}m_k(t)\delta[\tilde\gamma_k(t)- f_k({\gamma}_k(t-1))]
\delta[m_k(t)-\chi_k (\gamma_k(t)+\tilde{\gamma}_k(t))] \nonumber\\ &\times 
\delta\left[\tilde \gamma_k(t)-\lambda_km_k(t)+\sum_{s\leq t}\hat{\mathcal G}_k(t,s)m_k(s)-\phi_k(t) \right]{\mathrm e}^{ {\rm i}\sum_{t\leq 0}\gamma_k(t)l(t)}.\label{gfunctional}
\end{align}
Moreover, the entries of the respond matrices $\{\mathcal  G_k\}$ in \eqref{{MG}} are re-expressed in terms of the Gaussian processes $\{\phi_k(t)\}$  as
\begin{equation}
\mathcal G_k(t,s)=-\mathbb E\left[\frac{\partial m_k(t)}{\partial \phi_k(s)}\right]. \label{response} 
\end{equation}
\subsection{Vanishing memories}
The single-site generating functionals $Z_k(\{l(t)\})$ \eqref{gfunctional} (for $k=1,2$) refer to the stochastic processes
\begin{subequations}
	\label{effst}
	\begin{align}
	\tilde\gamma_k(t)&=f_k(\gamma_k(t-1))\\
	\gamma_k(t)&=\frac{1}{\lambda_k-\hat{\mathcal G}_k(t,t)}\left[\left(\frac{1}{\chi_k}+\hat{\mathcal G}_k(t,t)-\lambda_k\right)\tilde\gamma_k(t)+\sum_{s<t}\hat{\mathcal{G}}_k(t,s)(\gamma_k(s)+\tilde\gamma_k(s))    -\frac{\phi_k(t)}{\chi_k}\right].\label{gamma(t)}
	\end{align}	
\end{subequations} 
We next show that the conditions 
\begin{subequations}	
	\label{cond}
	\begin{align}
	\frac{\partial\gamma_k(t)}{\partial \phi_k(s)}=0\quad t>s\label{cond1}\\
	\mathbb E[\tanh'(h_k+\gamma_k(t))]=\chi_k,  \quad \forall t\label{cond2}
	\end{align}
\end{subequations}
are consistent with the stochastic processes \eqref{effst} and the uniqueness of \eqref{cond} follows inductively over discrete time. Specifically, from \eqref{cond} the entries of the response matrices \eqref{response} read as
\begin{align}
\mathcal G_k(t,s)&=-\chi_k\mathbb E\left[\frac{\partial (\gamma_k(t)+\tilde \gamma_k(t))}{\partial \phi_k(s)}\right]\\
&=\frac{\delta(t-s)}{\lambda_k-\hat{\mathcal G}_k(t,t)}-\chi_k\delta(t-1-s)\mathbb E\left[\frac{\partial \gamma_k(s)}{\partial \phi_k(s)}f_{k}'(\gamma_k(s))\right]\\
&=\frac{\delta(t-s)}{\lambda_k-\hat{\mathcal G}_k(t,t)}+\frac{\delta(t-1-s)}{\lambda_k-\hat{\mathcal G}_k(s,s)}\mathbb E\left[f_{k}'(\gamma_k(s))\right]\\
&=\frac{\delta(t-s)}{\lambda_k-\hat{\mathcal G}_k(t,t)}\label{keyghat}
\end{align}
where $\mathbb E\left[f_{k}'(\gamma_k(s))\right]=0$ follows from the condition \eqref{cond2}. Moreover, from \eqref{ghats}, the equation \eqref{keyghat} implies $\hat{\mathcal G}_k(t,s)=\hat{\mathcal G}_k(t,t)\delta(t-s)$. Actually, we have the explicit solutions
\begin{align}
\mathcal G_k(t,s)&=\chi_k\delta(t-s),\quad k=1,2\label{simplifed}\\
\mathcal{\hat G}_1(t,s)&=\chi_2 I'(\chi_1\chi_2)\delta(t-s)\\
\mathcal{\hat G}_2(t,s)&=\frac{\chi_1 I'(\chi_1\chi_2)}{\alpha}\delta(t-s).
\end{align}
These results lead \eqref{gamma(t)} to $\gamma_k(t)=-\phi_k(t)$ which shows the consistency of \eqref{cond1}. In summary, the single-site generating functionals read as
\begin{equation}
Z_k(\{l(t)\})\doteq \int  \{{\rm d}\gamma_k(t)\}\; \mathcal N(\gamma_k(0)\vert 0,\hat q_k)\mathcal N(\gamma_k(1),\dots,\gamma_k(T)\vert \matr 0, \mathcal C_{\gamma_k})\;{\mathrm e}^{ {\rm i}\sum_{t=0}^{T}\gamma_k(t)l(t)}
\end{equation}
where for convenience $\mathcal C_{\gamma_k}(t,s)\doteq \mathcal C_{\phi_k}(t,s)$. In the next section, we will give an explicit recursion of the two-time covariance matrices $\mathcal C_{\gamma_k}(t,s)$ from which the condition \eqref{cond2} follows (see \ref{variance terms}).   
\subsection{Computation of the two-time covariance matrices}
From \eqref{powerI} and \eqref{psi1}\&\eqref{frtrs} we respectively write 
\begin{align}
{I}''(\chi)&=\sum_{n=1}^{\infty} c_n (n-1)\chi^{n-2}\\
{\psi_1'}(\chi)&= I'(\chi)+\chi I''(\chi)\\
&=\sum_{n=1}^{\infty} c_n n\chi^{n-1}\label{psieq}
\end{align} 
Recall that $\mathcal  G_k(t,s)=\chi_k\delta(t,s)$ for $k=1,2$. Hence, we have from \eqref{covfs} that
\begin{align}
\left[\begin{array}{c}
\mathcal C_{\gamma_1}(t,s)\\
\mathcal C_{\gamma_2}(t,s)
\end{array}\right]=\left[\begin{array}{cc}
\chi_2^2 I''(\chi)& {\psi_1'}(\chi)\\
\frac{\psi_1'(\chi)}{\alpha}&\frac{\chi_1^2}{\alpha} I''(\chi)
\end{array}\right]~ \left[\begin{array}{c}
\mathcal C_{1}(t,s)\\\mathcal C_{2}(t,s)
\end{array}\right]. \label{takereplica}
\end{align}
Here, the entries of the order matrix $\mathcal C_{k}$ read as
\begin{equation}
\mathcal C_k(t,s)= \chi_k^2\mathbb E[(\gamma_k(t)+\tilde\gamma_k(t))(\gamma_k(s)+\tilde\gamma_k(s))]
\end{equation}
with $\tilde\gamma_k(t)\doteq f_k(\gamma_k(t-1))$. Moreover, we recall that the condition \eqref{cond2} (see \ref{variance terms} for the derivation) implies $\mathbb E[f_k'(\gamma_k(t))]=0,\forall t$. Thus, by Stein's Lemma we get 
\begin{equation}
\mathbb E[\gamma_k (t)\tilde\gamma_k(s)]=0,\forall t,s.
\end{equation}
This leads to
\begin{align}
\mathcal C_k(t,s)&=\chi_k^2(\mathcal C_{\gamma_k}(t,s)+\mathcal C_{\tilde\gamma_k}(t,s)) ~~\text{with}~~ \mathcal C_{\tilde\gamma_k}(t,s)\doteq \mathbb E[\tilde \gamma_k(t)\tilde\gamma(s)]. \label{relation}
\end{align}

From \eqref{takereplica} we write
\begin{align}
\left[\begin{array}{c}
\mathcal C_{1}(t,s)\\
\mathcal C_{2}(t,s)
\end{array}\right]=\frac{1}{D}
\left[\begin{array}{cc}
{\chi_1^2}{I}''(\chi)& -\alpha{\psi_1'(\chi)}\\
-{\psi_1'(\chi )}&\alpha\chi_2^2{ I}''(\chi)
\end{array}\right]\left[\begin{array}{c}
\mathcal C_{\gamma_1}(t,s)\\\mathcal C_{\gamma_2}(t,s)
\end{array}\right]\label{okay}
\end{align}
where for short we  have defined 
\begin{align}
D&\doteq(\chi{I}''(\chi)-{\psi_1'}(\chi)) (\chi{I}''(\chi)+{\psi_1'}(\chi))\\
&=-I'(\chi)(I'(\chi)+2\chi I''(\chi)).\label{deq}
\end{align} 
Plugging the relation \eqref{relation} in \eqref{okay} we get
\begin{align}
\left[\begin{array}{c}
\mathcal C_{\tilde\gamma_1}(t,s)\\
\mathcal C_{\tilde\gamma_2}(t,s)
\end{array}\right]=\frac{1}{D}
\left[\begin{array}{cc}
{{I}''}(\chi)-D& -\alpha\frac{\psi_1'(\chi)}{\chi_1^2}\\
-\frac{{\psi_1'(\chi)}}{\chi_2^2}& \alpha{{I}''(\chi)}-D
\end{array}\right]\left[\begin{array}{c}
\mathcal C_{\gamma_1}(t,s)\\ \mathcal C_{\gamma_2}(t,s)
\end{array}\right].
\end{align}
Then, we have obtained the desired expression
\begin{align}
\left[\begin{array}{c}
\mathcal C_{\gamma_1}(t,s)\\
\mathcal C_{\gamma_2}(t,s)
\end{array}\right]= 
\frac{1}{\alpha+ \chi^2D-(1+\alpha)\chi^2{I}''(\chi)}\left[\begin{array}{cc}
\chi^2(\alpha{I}''(\chi)-D)& \alpha\chi_2^2\psi_1'(\chi)\\
\chi_1^2{\psi_1'(\chi)}& \chi^2({I}''(\chi)-D)
\end{array}\right]
\left[\begin{array}{c}
\mathcal C_{\tilde\gamma_1}(t,s)\\
\mathcal C_{\tilde\gamma_2}(t,s)
\end{array}\right].\label{expres1}
\end{align}
\subsection{The property of fixed variances: $\mathcal C_{\gamma_k}(t,t)=\hat q_k$ for all $t$}\label{variance terms}
Note that $\mathcal C_{\gamma_k}(0,0)=\hat q_k$. We next show the consistency of the solution $C_{\gamma_k}(1,1)=\hat q_k$. Indeed, $C_{\gamma_k}(0,0)=C_{\gamma_k}(1,1)=\hat q_k$ yields $\mathcal C_k(1,1)=1-\chi_k$ and plugging this expression into \eqref{takereplica} leads $C_{\gamma_k}(1,1)$ immediately to the definitions of $\hat q_k$ in \eqref{s.order}.

We use the aforementioned consistency as a shortcut to show the relation (see \eqref{expres1})
\begin{align}
\left[\begin{array}{c}
\hat q_1\\
\hat q_2
\end{array}\right]= 
\frac{1}{\alpha+ \chi^2D-(1+\alpha)\chi^2{I}''(\chi)}\left[\begin{array}{cc}
\chi^2(\alpha{I}''(\chi)-D)& \alpha\chi_2^2\psi_1'(\chi)\\
\chi_1^2{\psi_1'(\chi)}& \chi^2({I}''(\chi)-D)
\end{array}\right]
\left[\begin{array}{c}
\frac{1-\chi_1}{\chi_1^2}-\hat q_1\\
\frac{1-\chi_2}{\chi_2^2}-\hat q_2
\end{array}\right]\label{g2}
\end{align}
where we note that $\mathcal C_{\gamma_k}(0,0)=\hat q_k$ yields the expression $\mathcal C_{\tilde\gamma_k}(1,1)=\frac{1-\chi_k}{\chi_k^2}-\hat q_k$. Thus, from \eqref{expres1} it follows inductively over discrete time that $\mathcal C_{\gamma_k}(t,t)=\hat q_k$ for all $t$.


\section{Derivations of \eqref{dergf} and \eqref{newfp}}\label{dcovex2}
It is easy to show that the limits in \eqref{dergf} can be expressed as
\begin{equation}
\left[\begin{array}{cc}
a_{11}&a_{12}\\
a_{21}&a_{22}
\end{array}\right]=\matr \Theta=-\left[\begin{array}{cc}
\frac{\psi_2^2}{\chi^2}{{\rm G}}_\matr W'(\lambda)+1
&  \frac{1}{\chi_1^2}(\lambda{\rm G}'_\matr W(\lambda)+{\rm G}_\matr W(\lambda))
\\
\frac{1}{\alpha\chi_2^2}(\lambda{\rm G}'_\matr W(\lambda)+{\rm G}_\matr W(\lambda))&\frac{\psi_1^2}{\chi^2}{{\rm G}}'_{\matr W^\top}(\lambda)+1
\end{array}\right]\label{covx}
\end{equation}
with $\lambda=\frac{\psi_1\psi_2}{\chi}$, $\chi=\chi_1\chi_2$ and ${\rm G}_\matr W'$ denoting the derivative of the Green function ${\rm G}_\matr W$ \eqref{Gf}. In the sequel, we will derive the result
\begin{equation}
\matr \Theta=\frac{1}{\alpha+ \chi^2D-(1+\alpha)\chi^2{I}''(\chi)}
\left[\begin{array}{cc}
\chi^2(\alpha{I}''(\chi)-D)& \alpha\chi_2^2\psi_1'(\chi)\\
\chi_1^2{\psi_1'(\chi)}& \chi^2({I}''(\chi)-D)
\end{array}\right]\label{showfgood}.
\end{equation}
Here, $\psi_1'(\chi)$ and $D$ are as in \eqref{psieq} and \eqref{deq}, specifically
\begin{equation*}
\psi_1'(\chi)=I'(\chi)+\chi I''(\chi) \quad \text{and} \quad D=-I'(\chi)(I'(\chi)+2\chi I''(\chi)).
\end{equation*}
Firstly, from \eqref{expres1} the result \eqref{showfgood} implies that the coefficients $\{a_{kk'}\}$ are those in~\eqref{covex1}. Secondly, from \eqref{g2} it implies the equations in \eqref{newfp}. Specifically, we have
\begin{equation}
\left[\begin{array}{c}
\hat q_1\\ \hat q_2
\end{array}\right]=\matr \Theta\left[\begin{array}{c}
\frac{1-\chi_1}{\chi_1^2}-\hat q_1
\\ \frac{1-\chi_2}{\chi_2^2}-\hat q_2
\end{array}\right] \iff \left[\begin{array}{c}
\hat q_1\\ \hat q_2
\end{array}\right]=({\bf I}+\matr \Theta)^{-1}\matr \Theta\left[\begin{array}{c}
\frac{1-\chi_1}{\chi_1^2}
\\ \frac{1-\chi_2}{\chi_2^2}
\end{array}\right].
\end{equation}
Moreover, from \eqref{good1} and \eqref{good2} we note that
\begin{equation}
\left[\begin{array}{c}
\psi_1\\ \psi_2
\end{array}\right]=\left[\begin{array}{c}
\frac{\chi}{{\rm G}_{\matr W^\top}(\lambda)}\\ \frac{\chi}{{\rm G}_\matr W(\lambda)}
\end{array}\right].\label{goodpsi}
\end{equation}

To show the diagonal terms in \eqref{showfgood}, we use the expressions \eqref{u1} and \eqref{u2} to write 
\begin{align}
\frac{1}{1-\frac{\chi^2}{\psi_2^2}{\rm R}'_{\matr W}(\frac \chi{\psi_2})}-1&=\frac{\chi^2(\alpha{I}''(\chi)-D)}{\alpha+ \chi^2D-(1+\alpha)\chi^2{I}''(\chi)}\label{r1}\\
\frac{1}{1-\frac{\chi^2}{\psi_1^2}{\rm R}'_{\matr W^\top}(\frac \chi{\psi_1})}-1&=\frac{\chi^2({I}''(\chi)-D)}{\alpha+ \chi^2D-(1+\alpha)\chi^2{I}''(\chi)}\label{r2}.
\end{align}
On the other hand, we have from \eqref{drtrs} and \eqref{goodpsi} 
\begin{align}
\frac{1}{1-\frac{\chi^2}{\psi_2^2}{\rm R}'_{\matr W}(\frac \chi{\psi_2})}&=-\frac{\psi_2^2}{\chi^2}{\rm G}'_{\matr W}(\lambda)\label{n1}\\
\frac{1}{1-\frac{\chi^2}{\psi_1^2}{\rm R}'_{\matr W^\top}(\frac \chi{\psi_1})}&=-\frac{\psi_1^2}{\chi^2}{\rm G}'_{\matr W^\top}(\lambda)\label{n2}.
\end{align}
This completes the derivation for the diagonal terms. 

Next we show the off-diagonals in \eqref{showfgood}. To this end we note that
\begin{align}
\tilde {\rm G}'_{\matr W}(\lambda)&\doteq\int \frac{t{\rm dP}_{\matr W}(t)}{(\lambda-t)^2}\\
&=-\frac{{\rm d}\lambda {\rm G}_{\matr W}(\lambda)}{{\rm d}\lambda}\\
&=-\lambda{\rm G}'_{\matr W}(\lambda)-{\rm G}_{\matr W}(\lambda).
\end{align}
Second, \eqref{good1} implies $\int\frac{{\rm dP}_{\matr W}(t)}{(\psi_1-\frac{\chi}{\psi_2}t)}=1$. Hence,  we have
\begin{align}
0&=\frac{{\rm d}}{{\rm  d}\chi}\int\frac{{\rm dP}_{\matr W}(t)}{(\psi_1-\frac{\chi}{\psi_2}t)}\\
&=-\psi_1'(\chi)\underbrace{\int\frac{{\rm dP}_{\matr W}(t)}{(\psi_1-\frac{\chi}{\psi_2}t)^2}}_{-\frac{\psi_2^2}{\chi^2}{\rm G}'_{\matr W}(\lambda)}+ (\psi_2-{\chi}\psi_2'(\chi))\underbrace{\frac{1}{\psi_2^2}\int\frac{t{\rm dP}_{\matr W}(t)}{(\psi_1-\frac{\chi}{\psi_2}t)^2}}_{\frac{1}{\chi^2}\tilde {\rm G}'_{\matr W}(\lambda)}.
\end{align}
From this, we write 
\begin{align}
\tilde {\rm G}'_{\matr W}(\lambda)&=\frac{\psi_1'(\chi)\chi^2}{[1-\frac{\chi^2}{\psi_2^2}{\rm R}'_{\matr W}(\frac \chi{\psi_2})][\psi_2-\chi\psi_2'(\chi)]}\\
&=\frac{\alpha\psi_1'(\chi)\chi^2}{[1-\frac{\chi^2}{\psi_2^2}{\rm R}'_{\matr W}(\frac \chi{\psi_2})][1 -\chi^2I''(\chi)]}\\
&=\frac{\alpha\psi_1'(\chi)\chi^2}{\alpha+ \chi^2D-(1+\alpha)\chi^2{I}''(\chi)}.
\end{align}
Here, from \eqref{r1} we use the relation
\begin{equation}
\frac{1}{1-\frac{\chi^2}{\psi_2^2}{\rm R}'_{\matr W}(\frac \chi{\psi_2})}=\frac{1 -\chi^2I''(\chi)}{\alpha+ \chi^2D-(1+\alpha)\chi^2{I}''(\chi)}.
\end{equation}
Hence, we complete the derivation of \eqref{showfgood}.


\section{Convergence analysis of the two-time covariances}\label{dervconv}
We introduce the functions
\begin{equation}
g_k(x)\doteq	\mathbb E[f_k(z_k)f_k(z_k')]
\end{equation} 
where $z_k$ and $z_k'$ are zero-mean Gaussian with variances $\hat q_k$ and covariance $x$. 
Hence, \eqref{covex1} reads as
\begin{equation}
\left[\begin{array}{c}
\mathcal C_{\gamma_1}(t,s)\\
\mathcal C_{\gamma_2}(t,s)
\end{array}\right]\doteq \left[\begin{array}{cc}
a_{11}&a_{12}\\
a_{21}&a_{22}
\end{array}\right]\left[\begin{array}{c}
g_1(\mathcal C_{\gamma_k}(t-1,s-1))\\
g_2(\mathcal C_{\gamma_k}(t-1,s-1))
\end{array}\right].\label{withg}
\end{equation} 
\begin{remark}\label{remgod}
The functions $g_k(x)$ for $k=1,2$ satisfy $0<g_k(0)$ and are strictly increasing on $[0,\hat q_k]$. Moreover, their derivatives are given by $g'_k(x)=\mathbb E[f'_k(z_k)f'_k(z_k')]$.
\end{remark}
The derivation of the remark is given at the end of this section. 

Assuming $\mathcal C_{\gamma_k}(t-1,s-1)\geq 0$, we write for $t\neq s$ and $k\neq k'$
\begin{align}
\mathcal C_{\gamma_k}(t,s)&=a_{kk}g_k(\mathcal C_{\gamma_k}(t-1,s-1))+a_{kk'}g_{k'}(\mathcal C_{\gamma_{k'}}(t-1,s-1))\\
&<a_{kk}g_k(\mathcal C_{\gamma_k}(t,s))+a_{kk'}g_{k'}(\mathcal C_{\gamma_{k'}}(t,s))=\mathcal C_{\gamma_k}(t+1,s+1)\\
&<a_{kk}g_k(\hat q_k)+a_{kk'}g_{k'}(\hat q_{k'})=\hat q_k.
\end{align}
Here the strict inequalities follow from the fact that $a_{kk'}>0$ for $k\neq k'$,$a_{kk}\geq0$ and $g_k(x)$ is strictly increasing on $[0,\hat q_k]$ with $0<g_k(0)$. Since $\mathcal C_{\gamma_k}(t,0)=0,\forall t>0$, it then follow inductively over iteration steps that 
\begin{equation}
\mathcal C_{\gamma_k}(t-1,s-1)<\mathcal C_{\gamma_k}(t,s)<\hat q_k,\quad \forall t\neq s.\label{nstrictp}
\end{equation}

Recall that $\Delta_{\gamma_k}(t,s)=2(\hat q_k-\mathcal C_{\gamma_k}(t,s))$. Hence, linearizing \eqref{withg} around the stable solutions gives
\begin{equation}
\left[\begin{array}{c}
\Delta_{\gamma_k}(t,s)\\
\Delta_{\gamma_1}(t,s)
\end{array}\right]\simeq \left[\begin{array}{cc}
g'_1a_{11}&g'_2a_{12}\\
g'_1a_{21}&g'_2a_{22}
\end{array}\right]
\left[\begin{array}{c}
\Delta_{\gamma_k}(t-1,s-1)\\
\Delta_{\gamma_k}(t-1,s-1)
\end{array}\right]\label{withl}
\end{equation}
It is easy to show that the absolute value of the maximum eigenvalue of the $2\times 2$ Jacobian matrix in \eqref{withl} reads as
\begin{equation}
\mu_\gamma=\frac{1}{2}(g'_1a_{11}+g'_2a_{22})+\frac 1 2\sqrt{(g'_1a_{11}-g'_2a_{22})^2+4g'_1g'_2a_{12}a_{21}}.
\end{equation}
Hence, we have the limits
\begin{equation}
\lim_{t,s\to \infty}\Delta_{\gamma_k}(t,s)=0
\end{equation}
if and only if $\mu_\gamma<1$ (else, we have $\lim_{t,s\to \infty}\Delta(t,s)>0$). The asymptotic decay of the error is dominated by the largest eigenvalue and the convergence rate
is given by
\begin{equation}
\lim_{t,s\to \infty}\frac{\Delta_{\gamma_k}(t+1,s+1)}{\Delta_{\gamma_k}(t,s)}=\mu_\gamma, \quad k=1,2.
\end{equation}

Next we show that the stability condition of the TAP equations \eqref{AT} becomes necessary for the bound $\mu_\gamma<1$. Firstly,  $\mu_\gamma<1$ implies that
\begin{equation}
\max(g'_1a_{11},g'_2a_{22})=\frac{1}{2}(g'_1a_{11}+g'_2a_{22})+\frac 1 2\sqrt{(g'_1a_{11}-g'_2a_{22})^2}<1. \label{bounda}
\end{equation} 
Furthermore, it is immediate to show that 
\begin{equation}
g'_k=\frac{1}{\chi_k^2}\mathbb E[(\tanh'(h_k+\sqrt{\hat q_k}u))^2]-1.
\end{equation}
Then, it turns out that (see \eqref{n1} and \eqref{n2})
\begin{equation}
g'_ka_{kk}=\frac{{\rm R}'_k\mathbb E[(\tanh'(h_k+\sqrt{\hat q_k}u))^2]-\chi_k^2{\rm R}_k'}{1-\chi_k^2{\rm R}_k'}.
\end{equation}
Hence, \eqref{bounda} holds if and only if
\begin{equation}
{\rm R}_k'\mathbb E[(\tanh'(h_k+\sqrt{\hat q_k}u))^2]<1,\quad  k=1,2.
\end{equation}
\subsection{Derivation of Remark~\ref{remgod}}
By the representation of the Gaussian density in terms of its characteristic function we~have
\begin{equation}
g_k(x)=\frac{1}{(2\pi)^2}\int {\rm d}z{\rm d}z'{\rm d}y{\rm d}y'\;f_k(z) f_k(z') {\mathrm e}^{-{\rm i}(yz+y' z') -\frac{\hat q_k}{2}(y^2+y'^2)}{\mathrm e}^{-xyy'}\label{Gcar}.
\end{equation}
Thus, we have the derivative of $g_k(x)$ as $g'_k(x)=\mathbb E[f'_k(z_k)f'_k(z_k')]$. We now recall the following useful result from \cite{CakmakOpper19}.
\begin{remark}\cite{CakmakOpper19}\label{useful}
	Let $z$ and $z'$ be Gaussian random variables and be identically distributed. Furthermore, let the covariance between $z$ and $z'$ be positive. Moreover, let the function $f$ have derivatives of all orders in~$\RR$. Then, the covariance between the random variables $f(z)$ and $f(z')$ is positive, too. 
\end{remark}
This result implies that $g_k(x)>0$, $g'_k(x)>0$ on $x\in (0,\hat q_k]$.  Moreover, since $h_k\neq 0$, we have the positivity for the boundary case $x=0$, i.e.  $g_k(0)>0$, $g'_k(0)>0$. This completes the derivation of Remark~\ref{remgod}.


\bibliographystyle{iopart-num}
\bibliography{mybib}
\end{document}